\documentclass{article} 
\usepackage{iclr2019_conference,times}

\usepackage[utf8]{inputenc} 
\usepackage[T1]{fontenc}    
\usepackage{hyperref}       
\usepackage{url}            
\usepackage{booktabs}       
\usepackage{amsfonts}       
\usepackage{nicefrac}       
\usepackage{microtype}      
\usepackage{adjustbox}
\usepackage{graphbox}
\usepackage{graphicx}
\usepackage{subfigure}
\usepackage{booktabs} 
\usepackage{array}
\usepackage{url}
\usepackage{amsmath, nccmath}
\usepackage{bm}
\usepackage{xcolor}
\usepackage{wrapfig}

\usepackage{colortbl}

\usepackage[]{algorithm2e}
\usepackage{algorithmic}

\usepackage{hyperref}

\usepackage{bbm}

\newcommand{\lowRR}{\operatorname{LowRR}}

\newcommand{\pd}[2]{\frac{\partial #1}{\partial #2}}

\newcommand{\pp}[1]{\left( #1 \right)}

\newcommand{\argmin}{\operatornamewithlimits{argmin}}

\newcommand{\expect}[2]{\mathbbm{E}_{#1}\left[ #2 \right]}
\newcommand{\R}[1]{\mathbbm{R}^{#1}}

\newcommand{\ConvBatch}{\operatorname{ConvBatch}}
\newcommand{\ConvUnit}{\operatorname{ConvUnit}}
\newcommand{\BatchNorm}{\operatorname{BatchNorm}}
\newcommand{\ReLU}{\operatorname{ReLU}}

\title{Meta-Learning Update Rules for Unsupervised Representation Learning}

\author{%
  Luke Metz \\
  Google Brain\\
  \texttt{lmetz@google.com} \\
  \And
  Niru Maheswaranathan \\
  Google Brain\\
  \texttt{nirum@google.com} \\
  \And
  Brian Cheung \\
  University of California, Berkeley \\
  \texttt{bcheung@berkeley.edu} \\
  \And
  Jascha Sohl-Dickstein \\
  Google Brain \\
  \texttt{jaschasd@google.com} \\
}

\begin{document}
\iclrfinalcopy
\maketitle

\begin{abstract}
A major goal of unsupervised learning is to discover data representations that are useful for subsequent tasks, without access to supervised labels during training. Typically, this involves minimizing a surrogate objective, such as the negative log likelihood of a generative model, with the hope that representations useful for subsequent tasks will arise as a side effect. 
In this work, we propose instead to directly target later desired tasks by meta-learning an unsupervised learning rule which leads to representations useful for those tasks. 

Specifically, we target semi-supervised classification performance, and we meta-learn an algorithm -- an unsupervised weight update rule -- that produces representations useful for this task. 
Additionally, we constrain our unsupervised update rule to a be a biologically-motivated, neuron-local function, which enables it to generalize to different neural network architectures, datasets, and data modalities.
We show that the meta-learned update rule produces useful features and sometimes outperforms existing unsupervised learning techniques. We further show that the meta-learned unsupervised update rule generalizes to train networks with different widths, depths, and nonlinearities. It also generalizes to train on data with randomly permuted input dimensions and even generalizes from image datasets to a text task.
\end{abstract}

\section{Introduction} \label{introduction}
Supervised learning has proven extremely effective for many problems where large amounts of labeled training data are available.
There is a common hope that unsupervised learning will prove similarly powerful in situations where labels are expensive, impractical to collect, or where the prediction target is unknown during training.
Unsupervised learning however has yet to fulfill this promise. 
One explanation for this failure is that unsupervised representation learning algorithms are typically mismatched to the target task.
Ideally, learned representations should linearly expose high level attributes of data (e.g. object identity) and perform well in semi-supervised settings. 
Many current unsupervised objectives, however, optimize for 
objectives such as log-likelihood of a generative model or reconstruction error, producing useful representations only as a side effect.

Unsupervised representation learning seems uniquely suited for meta-learning
\citep{hochreiter2001learning, schmidhuber1995learning}. 
Unlike most tasks where meta-learning is applied, unsupervised 
learning does not define an explicit objective, which makes it impossible to phrase the task as a standard optimization problem.
It is possible, however, to directly express a {\em meta-objective} that captures the quality of representations produced by an unsupervised update rule by evaluating the usefulness of the representation for candidate tasks. 
In this work, we propose to \emph{meta-learn} an unsupervised update rule by meta-training on a meta-objective that directly optimizes the utility of the unsupervised representation. Unlike hand-designed unsupervised learning rules, this meta-objective directly targets the usefulness of a representation generated from unlabeled data for later supervised tasks.

By recasting unsupervised representation learning as meta-learning, we treat the creation of the unsupervised update rule as a transfer learning problem.
Instead of learning transferable features, we learn a transferable learning rule which does not require access to labels and generalizes across both data domains and neural network architectures.
Although we focus on the meta-objective of semi-supervised classification here, in principle a learning rule could be optimized to generate representations for \textit{any} subsequent task.

\section{Related Work}

\subsection{Unsupervised Representation Learning} \label{related:unsupervised}

Unsupervised learning is a topic of broad and diverse interest. Here we briefly review several techniques that can lead to a useful latent representation of a dataset. In contrast to our work, each method imposes a manually defined training algorithm or loss function whereas we \emph{learn} the algorithm that creates useful representations as determined by a meta-objective.

Autoencoders \citep{hinton2006reducing} work by first compressing and optimizing reconstruction loss. Extensions have been made to de-noise data  \citep{vincent2008extracting, vincent2010stacked}, as well as compress information in an information theoretic way \citep{kingma2013auto}. \citet{le2011building} further explored scaling up these unsupervised methods to large image datasets.

Generative adversarial networks \citep{goodfellow2014generative} take another approach to unsupervised feature learning. Instead of a loss function, an explicit min-max optimization is defined to learn a generative model of a data distribution. Recent work has shown that this training procedure can learn unsupervised features useful for few shot learning \citep{radford2015unsupervised, donahue2016adversarial, dumoulin2016adversarially}.

Other techniques rely on self-supervision where labels are easily generated to create a non-trivial `supervised' loss. Domain knowledge of the input is often necessary to define these losses. \citet{noroozi2016unsupervised} use unscrambling jigsaw-like crops of an image. Techniques used by \citet{misra2016shuffle} and \citet{sermanet2017time} rely on using temporal ordering from videos.

Another approach to unsupervised learning relies on feature space design such as clustering. \citet{coates2012learning} showed that k-means can be used for feature learning.  \citet{xie2016unsupervised} jointly learn features and cluster assignments. \cite{bojanowski2017unsupervised} develop a scalable technique to cluster by predicting noise. 
Other techniques such as \citet{schmidhuber1992learning}, \citet{hochreiter1999feature}, and \citet{olshausen1997sparse} define various desirable properties about the latent representation of the input, such as predictability, complexity of encoding mapping, independence, or sparsity, and optimize to achieve these properties.

\subsection{Meta Learning}

Most meta-learning algorithms consist of two levels of learning, or `loops' of computation: an \emph{inner loop}, where some form of learning occurs (e.g. an optimization process), and an \emph{outer loop} or {\em meta-training} loop, which optimizes some aspect of the inner loop, parameterized by {\em meta-parameters}. 
The performance of the inner loop computation for a given set of meta-parameters is quantified by a {\em meta-objective}.
{\em Meta-training} is then the process of adjusting the meta-parameters so that the inner loop performs well on this meta-objective.
Meta-learning approaches differ by the computation performed in the inner loop, the domain, the choice of meta-parameters, and the method of optimizing the outer loop.

Some of the earliest work in meta-learning includes work by \citet{schmidhuber1987evolutionary}, which explores a variety of meta-learning and self-referential algorithms. Similarly to our algorithm, \citet{bengio1990learning, bengio1992optimization} propose to learn a neuron local learning rule, though their approach differs in task and problem formulation. 
Additionally, \citet{runarsson2000evolution} meta-learn supervised learning rules which mix local and global network information. 
A number of papers propose meta-learning for few shot learning \citep{Vinyals2016,ravi2016optimization,mishra2017meta,finn17,snell2017prototypical}, though these do not take advantage of unlabeled data. Others make use of both labeled and unlabeld data \citep{ren2018meta}. \citet{hsu2018unsupervised} uses a task created with no supervision to then train few-shot detectors. \citet{garg2017supervising} use meta-learning for unsupervised learning, primarily in the context of clustering and with a small number of meta-parameters. 

To allow easy comparison against other existing approaches, we present a more extensive survey of previous work in meta-learning in table form in Table \ref{tab meta compare}, highlighting differences in choice of task, structure of the meta-learning problem, choice of meta-architecture, and choice of domain.

To our knowledge, we are the first meta-learning approach to tackle the problem of unsupervised representation learning, where the inner loop consists of unsupervised learning. This contrasts with transfer learning, where a neural network is instead trained on a similar dataset, and then fine tuned or otherwise post-processed on the target dataset. 
We additionally believe we are the first representation meta-learning approach to generalize across input data modalities as well as datasets, the first to generalize across permutation of the input dimensions, and the first to generalize across neural network architectures (e.g. layer width, network depth, activation function).

\begin{table}[h!]
\footnotesize
\makebox[\textwidth][c]{
\begin{tabular}{ 
| >{\raggedright}p{4.3cm} 
| >{\raggedright}p{3.7cm} 
| >{\raggedright}p{2.0cm} 
!{\color{lightgray}\vrule} >{\raggedright}p{1.6cm} 
!{\color{lightgray}\vrule} >{\raggedright}p{1.4cm} 
| >{\raggedright\arraybackslash}p{2cm}
|}
\hline
&&\multicolumn{3}{c|}{}&\\[-0.35cm]
\noindent\parbox[b]{\hsize}{\center Method} & 
\noindent\parbox[b]{\hsize}{\center Inner loop updates} &
\multicolumn{3}{c|}{Outer loop updates, meta-}&
\noindent\parbox[b]{\hsize}{\center Generalizes to}\\
&&\multicolumn{3}{c|}{}&\\[-0.35cm]
&&\noindent\parbox[b]{\hsize}{\center parameters} & 
\noindent\parbox[b]{\hsize}{\center objective} & 
\noindent\parbox[b]{\hsize}{\center optimizer} &
 \\ 
&&&&&\\[-0.35cm]
\hline
&&&&&\\[-0.35cm]
Hyper parameter optimization \cite{jones2001taxonomy,snoek2012practical,bergstra2011algorithms,bergstra2012random} & many steps of optimization
& 
optimization hyper-parameters & training or validation set loss & Baysian methods, random search, etc & test data from a fixed dataset \\
&&&&&\\[-0.35cm]
\hline
&&&&&\\[-0.35cm]
Neural architecture search \cite{stanley2002evolving, Zoph2017, baker2016designing, zoph2017learning, real2017large} & supervised SGD training using meta-learned architecture & architecture & validation set loss & RL or evolution & test loss within similar datasets \\
&&&&&\\[-0.35cm]
\hline
&&&&&\\[-0.35cm]
Task-specific optimizer (eg for quadratic function identification) \citep{hochreiter2001learning} & adjustment of model weights by an LSTM & LSTM weights & task loss & SGD & similar domain tasks \\
&&&&&\\[-0.35cm]
\hline
&&&&&\\[-0.35cm]
Learned optimizers \cite{jones2001taxonomy,maclaurin2015gradient,andrychowicz2016learning,chen2016learning,li2016learning,wichrowska2017learned, Bello17}& many steps of optimization of a fixed loss function & parametric optimizer & average or final loss & SGD or RL & new loss functions (mixed success)\\
&&&&&\\[-0.35cm]
\hline
&&&&&\\[-0.35cm]
Prototypical networks \cite{snell2017prototypical} & apply a feature extractor to a batch of data and use soft nearest neighbors to compute class probabilities & weights of the feature extractor & few shot performance & SGD & new image classes within similar dataset \\
&&&&&\\[-0.35cm]
\hline
&&&&&\\[-0.35cm]
MAML \cite{finn17} & one step of SGD on training loss starting from a meta-learned network& initial weights of neural network & reward or training loss & SGD & new goals, similar task regimes with same input domain \\
&&&&&\\[-0.3cm]
\hline
&&&&&\\[-0.35cm]
Evolved Policy Gradient \cite{houthooft2018evolved} & performing gradient descent on a learned loss & parameters of a learned loss function & reward & Evolutionary Strategies & new environment configurations, both in and not in meta-training distribution. \\
&&&&&\\[-0.35cm]
\hline
&&&&&\\[-0.35cm]
Few shot learning \citep{Vinyals2016, ravi2016optimization, mishra2017meta} & application of a recurrent model, e.g. LSTM, Wavenet. & recurrent model weights & test loss on training tasks & SGD & new image classes within similar dataset. \\
&&&&&\\[-0.35cm]
\hline
&&&&&\\[-0.35cm]
Meta-unsupervised learning for clustering \cite{garg2017supervising} & run clustering algorithm or evaluate binary similarity function & clustering algorithm + hyperparameters, binary similarity function & empirical risk minimization & varied & new clustering or similarity measurement tasks \\
&&&&&\\[-0.35cm]
\hline
&&&&&\\[-0.35cm]
Learning synaptic learning rules \citep{bengio1990learning, bengio1992optimization} & run a synapse-local learning rule & parametric learning rule & supervised loss, or similarity to biologically-motivated network & gradient descent, simulated annealing, genetic algorithms & similar domain tasks \\
&&&&&\\[-0.35cm]
\hline
&&&&&\\[-0.35cm]
\textbf{Our work --- metalearning for unsupervised representation learning} & 
\textbf{many applications of an unsupervised update rule }
& \textbf{parametric update rule}
& \textbf{few shot classification after unsupervised pre-training} &
\textbf{SGD} &
\textbf{new base models (width, depth, nonlinearity), new datasets, new data modalities} \\
&&&&&\\
\hline
\end{tabular}
}
\vspace{0.1in}
\caption{A comparison of published meta-learning approaches.
\label{tab meta compare}}
\end{table}

{\parfillskip0pt\par}
\begin{figure}
\centering
\includegraphics[width=0.46\textwidth,align=c]{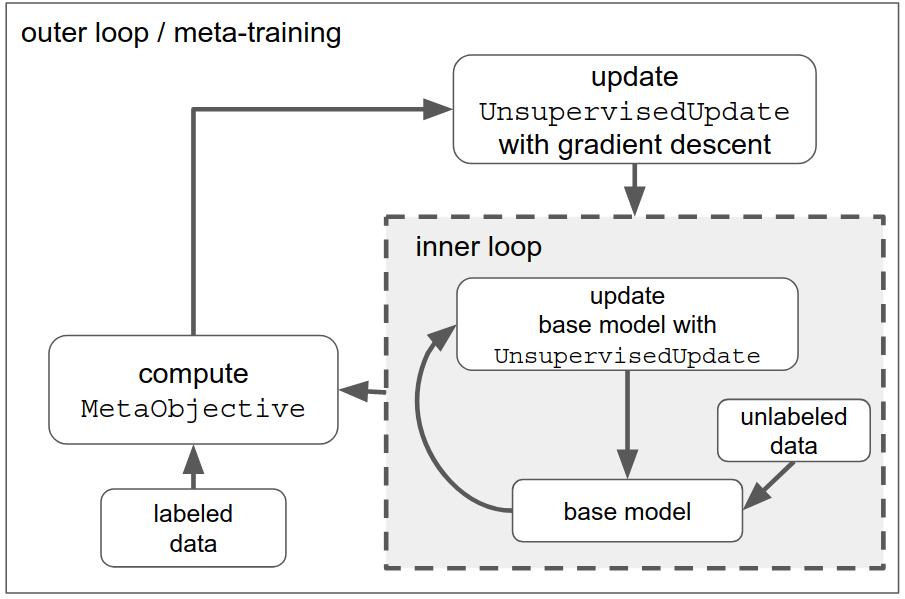}
\hspace{.1cm}
\includegraphics[width=0.52\textwidth,align=c]{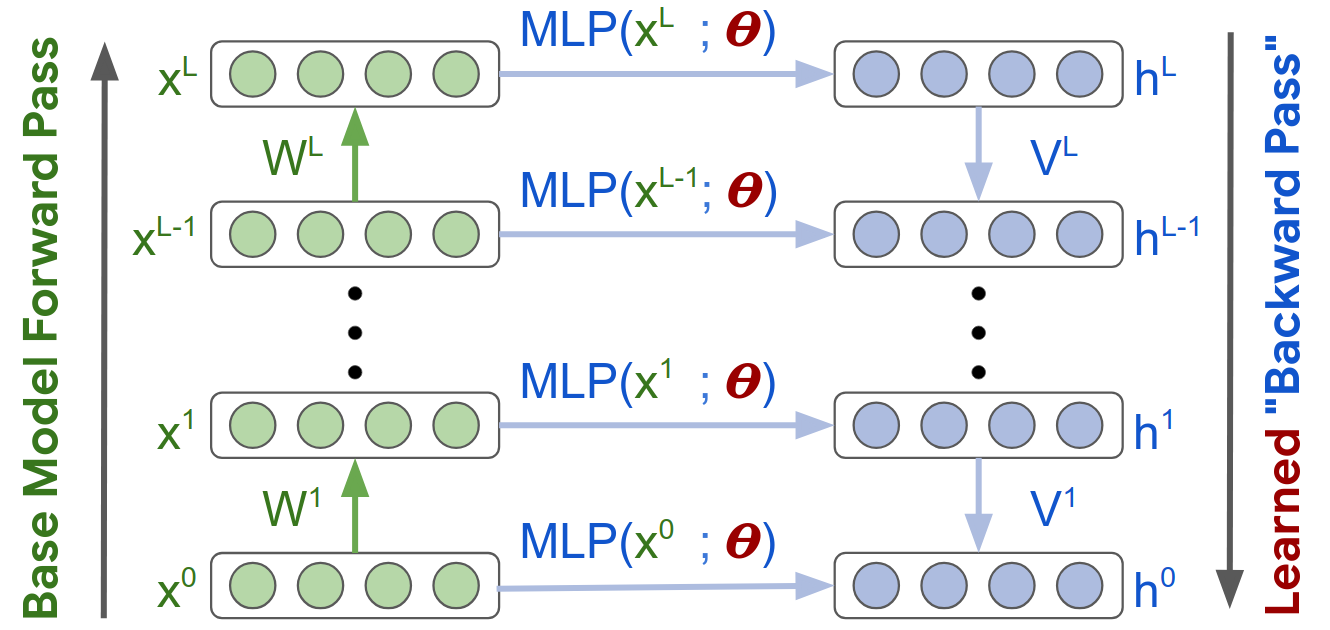}
\vspace{-5pt}
\caption{\textbf{Left:} Schematic for meta-learning an unsupervised learning algorithm. The inner loop computation consists of iteratively applying the $\text{UnsupervisedUpdate}$ to a base model. During meta-training the $\text{UnsupervisedUpdate}$ (parameterized by $\theta$) is itself updated by gradient descent on the $\text{MetaObjective}$. \textbf{Right:}
Schematic of the base model and $\text{UnsupervisedUpdate}$.
Unlabeled input data, $x_0$, is passed through the base model, which is parameterised by $W$ and colored green. The goal of the $\text{UnsupervisedUpdate}$ is to modify $W$ to achieve a top layer representation $x^L$ which performs well at few-shot learning. 
In order to train the base model, information is propagated backwards by the $\text{UnsupervisedUpdate}$ in a manner analogous to backprop.
Unlike in backprop however, the backward weights $V$ are decoupled from the forward weights $W$.
Additionally, unlike backprop, there is no explicit error signal as there is no loss.
Instead at each layer, and for each neuron, a learning signal is injected by a meta-learned MLP parameterized by $\theta$, with hidden state $h$. 
Weight updates are again analogous to those in backprop, and depend on the hidden state of the pre- and post- synaptic neurons for each weight.
\label{fig:schematic}
}
\end{figure}
\noindent

\section{Model Design}
We consider a multilayer perceptron (MLP) with parameters $\phi_t$ as the \textit{base model}. The inner loop of our meta-learning process trains this base model via iterative application of our learned update rule.
See Figure \ref{fig:schematic} for a schematic illustration and Appendix \ref{app:detail_schematic} for a more detailed diagram. 

In standard supervised learning, the `learned' optimizer is stochastic gradient descent (SGD). A supervised loss $l\left(x, y \right)$ is associated with this model, where $x$ is a minibatch of inputs, and $y$ are the corresponding labels.
The parameters $\phi_t$ of the base model are then updated iteratively
by performing SGD using the gradient $\frac{\partial l(x,y)}{\partial \phi_t}$. This supervised update rule can be written as 
$\phi_{t+1} = \text{SupervisedUpdate}(\phi_t, x_t, y_t; \theta)$, where $t$ denotes the inner-loop iteration or step.
Here $\theta$ are the meta-parameters of the optimizer, which consist of hyper-parameters such as learning rate and momentum.

In this work, our learned update is a parametric function which does not depend on label information,
$\phi_{t+1} = \text{UnsupervisedUpdate}(\phi_t, x_t; \theta)$.
This form of the update rule is general, it encompasses many unsupervised learning algorithms and all methods in Section \ref{related:unsupervised}.

In traditional learning algorithms, expert knowledge or a simple hyper-parameter search determines $\theta$, which consists of a handful of meta-parameters such as learning rate and regularization constants. In contrast, our update rule will have 
orders of magnitude more meta-parameters, including the weights of a neural network. We train these meta-parameters by performing SGD on the sum of the MetaObjective over the course of (inner loop) training in order to find optimal parameters $\theta^*$,

\begin{align}
    \mathcal{\theta}^* = \argmin_{\bm{\theta}} \expect{\text{task}}{\sum_t \text{MetaObjective}(\phi_t )}
    ,
\end{align}

that minimize the meta-objective over a distribution of training tasks. Note that $\phi_t$ is a function of $\theta$ since $\theta$ affects the optimization trajectory.

In the following sections, we briefly review the main components of this model: the base model, the $\text{UnsupervisedUpdate}$, and the $\text{MetaObjective}$.
See the Appendix for a complete specification.
Additionally, code and meta-trained parameters $\theta$ for our meta-learned $\text{UnsupervisedUpdate}$ is available\footnote{\url{https://github.com/tensorflow/models/tree/master/research/learning_unsupervised_learning}}.

\subsection{Base model}
Our base model consists of a standard fully connected multi-layer perceptron (MLP), with batch normalization \citep{Ioffe15}, and ReLU nonlinearities.
We chose this as opposed to a convolutional model to limit the inductive bias of convolutions in favor of learned behavior from the $\text{UnsupervisedUpdate}$.
We call the pre-nonlinearity activations $z^1,\cdots,z^L$, and post-nonlinearity activations $x^0,\cdots,x^L$, where $L$ is the total number of layers, and $x^0 \equiv x$ is the network input (raw data). 
The parameters are $\phi = \left\{W^1, b^1, V^1, \cdots, W^L, b^L, V^L \right\}$, where 
$W^l$ and $b^l$ are the weights and biases (applied after batch norm) for layer $l$, and $V^l$ are the corresponding weights used in the backward pass.

\subsection{Learned update rule}\label{sec UnsupervisedUpdate}
We wish for our update rule to generalize across architectures with different widths, depths, or even network topologies. 
To achieve this, we design our update rule to be {\em neuron-local}, so that updates are a function of pre- and post- synaptic neurons in the base model, and are defined for any base model architecture. This has the added benefit that it makes the weight updates more similar to synaptic updates in biological neurons, which depend almost exclusively on local pre- and post-synaptic neuronal activity \citep{whittington2017approximation}. In practice, we relax this constraint and incorporate some cross neuron information to decorrelate neurons (see Appendix \ref{sec app weight updates} for more information).

To build these updates, each neuron $i$ in every layer $l$ in the base model has an MLP, referred to as an update network, associated with it, with output $h^l_bi = \operatorname{MLP}\left(x^l_bi, z_bi^l, V^{l+1}, \delta^{l+1}; \theta \right)$ where $b$ indexes the training minibatch. The inputs to the MLP are the feedforward activations ($x^l$ \& $z^l$) defined above, and feedback weights and an error signal ($V^{l}$ and $\delta^{l}$, respectively) which are defined below.

All update networks share meta-parameters $\theta$.
Evaluating the statistics of unit activation over a batch of data has proven helpful in supervised learning \citep{Ioffe15}.
It has similarly proven helpful in hand-designed unsupervised learning rules, such as sparse coding and clustering.
We therefore allow $h^l_{bi}$ to accumulate statistics across examples in each training minibatch.

During an unsupervised training step, the base model is first run in a standard feed-forward fashion, populating $x^l_{bi},z^l_{bi}$.
As in supervised learning, an error signal $\delta^l_{bi}$ is then propagated backwards through the network. 
Unlike in supervised backprop, however, this error signal is generated by the corresponding update network for each unit. It is read out by linear projection of the per-neuron hidden state $h$, $\delta^l_{bi} = \operatorname{lin} \left(h^l_{bi}\right)$, and propogated backward using a set of learned `backward weights' $(V^l)^T$, rather than the transpose of the forward weights $(W^l)^T$ as would be the case in backprop (diagrammed in Figure~\ref{fig:schematic}). This is done to be more biologically plausible \citep{lillicrap2016random}.

Again as in supervised learning, the weight updates ($\Delta W^l$) are a product of pre- and post-synaptic signals.
Unlike in supervised learning however, these signals are generated using the per-neuron update networks: $\Delta W_{ij}^l = \operatorname{func}\left(h^l_{bi}, h^{l-1}_{bj}, W_{ij}\right)$. The full weight update (which involves normalization and decorrelation across neurons) is defined in Appendix \ref{sec app weight updates}.

\subsection{Meta-objective}
The meta-objective determines the quality of the unsupervised representations. In order to meta-train via SGD, this loss must be differentiable. 
The meta-objective we use in this work is based on fitting a linear regression to labeled examples with a small number of data points.  In order to encourage the learning of features that generalize well, we estimate the linear regression weights on one minibatch $\{x_a, y_a\}$ of $K$ data points, and evaluate the classification performance on a second minibatch $\{x_b, y_b\}$ also with $K$ datapoints,
\begin{fleqn}
\setlength{\abovedisplayskip}{4pt}
\setlength{\belowdisplayskip}{4pt}
\begin{align}
    &\hat{v} = \argmin_v \left(\left\|
        y_a - v^T x^L_a
    \right\|^2 + \lambda \left\| v \right\|^2\right)
    ,
    &\text{MetaObjective}(\cdot; \phi) = \operatorname{CosDist}\left( y_b, \hat{v}^T x^L_b \right)
    ,
\end{align}
\end{fleqn}
where $x^L_a$, $x^L_b$ are features extracted from the base model on data $x_a$, $x_b$, respectively. The target labels $y_a$, $y_b$ consist of one hot encoded labels and potentially also regression targets from data augmentation (e.g. rotation angle, see Section \ref{sec:sampling_data}). We found that using a cosine distance, $\operatorname{CosDist}$, rather than unnormalized squared error improved stability. Note this meta-objective is \textit{only} used during meta-training and \textit{not} used when applying the learned update rule. The inner loop computation is performed without labels via the $\text{UnsupervisedUpdate}$.

\section{Training the Update Rule}
\subsection{Approximate Gradient Based Training}
We choose to meta-optimize via SGD as opposed to reinforcement learning or other black box methods, due to the superior convergence properties of SGD in high dimensions, and the high dimensional nature of $\theta$. 
Training and computing derivatives through long recurrent computation of this form is notoriously difficult \citep{pascanu2013difficulty}. To improve stability and reduce the computational cost we approximate the gradients $\frac{\partial [\text{MetaObjective}]}{\partial \theta}$ via truncated backprop through time \citep{shaban2018truncated}. 
Many additional design choices were also crucial to achieving 
stability and convergence in meta-learning, including the use of batch norm, and restricting the norm of the $\text{UnsupervisedUpdate}$ update step (a full discussion of these and other choices is in Appendix \ref{app stability}).

\subsection{Meta-training Distribution and Generalization} \label{sec:sampling_data}
Generalization in our learned optimizer comes from both the form of the $\text{UnsupervisedUpdate}$ (Section \ref{sec UnsupervisedUpdate}), and from the meta-training distribution.  Our meta-training distribution is composed of both datasets and base model architectures.

We construct a set of training tasks consisting of CIFAR10 \citep{krizhevsky2009learning} and multi-class classification from subsets of classes from Imagenet \citep{ILSVRC15} as well as from a dataset consisting of rendered fonts (Appendix \ref{app:experimental:data}). We find that increased training dataset variation actually improves the meta-optimization process. To reduce computation we restrict the input data to 16x16 pixels or less during meta-training, and resize all datasets accordingly. For evaluation, we use MNIST \citep{lecun1998gradient}, Fashion MNIST \citep{xiao2017/online}, IMDB \citep{imdb}, and a hold-out set of Imagenet classes. We additionally sample the base model architecture. We sample number of layers uniformly between 2-5 and the number of units per layer logarithmically between 64 to 512.

As part of preprocessing, we permute all inputs along the feature dimension, so that the $\text{UnsupervisedUpdate}$ must learn a permutation invariant learning rule. Unlike other work, we focus explicitly on learning a learning algorithm as opposed to the discovery of fixed feature extractors that generalize across similar tasks. This makes the learning task \emph{much} harder, as the $\text{UnsupervisedUpdate}$ has to discover the relationship between pixels based solely on their joint statistics, and cannot ``cheat'' and memorize pixel identity.
To provide further dataset variation, we additionally augment the data with shifts, rotations, and noise. We add these augmentation coefficients as additional regression targets for the meta-objective--e.g. rotate the image and predict the rotation angle as well as the image class. For additional details, see Appendix \ref{app:experimental:data}.

\subsection{Distributed Implementation}

We implement the above models in distributed TensorFlow \citep{abadi2016tensorflow}. 
Training uses 512 workers, each of which performs a sequence of partial unrolls
of the inner loop $\text{UnsupervisedUpdate}$,
and computes gradients of the meta-objective asynchronously. 
Training takes $\sim$8 days, and consists of $\sim$200 thousand updates to $\theta$ with minibatch size 256. Additional details are in Appendix \ref{sec distributed implementation}.

\section{Experimental Results}
First, we examine limitations of existing unsupervised and meta learning methods. Then, we show meta-training and generalization properties of our learned optimizer and finally we conclude by visualizing how our learned update rule works. For details of the experimental setup, see Appendix \ref{app:experimental}.

\subsection{Objective Function Mismatch and Existing Meta-Learning Methods}
\begin{figure}[h]
\setlength{\belowcaptionskip}{-5pt}

\centering

\includegraphics[width=0.60\textwidth]{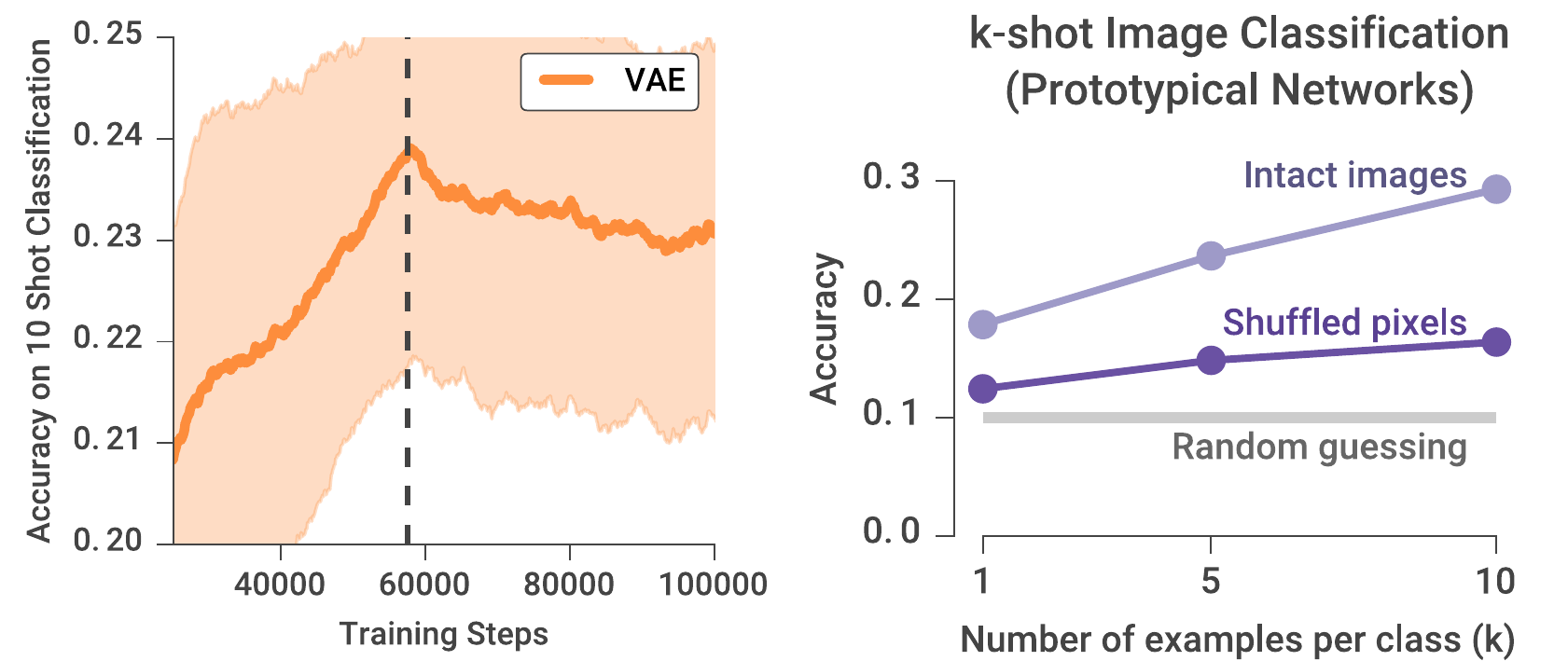}
\vspace{-10pt}
\caption{\textbf{Left:} Standard unsupervised learning approaches suffer from objective function missmatch. Continuing to optimize a variational auto-encoder (VAE) hurts few-shot accuracy after some number of steps (dashed line). \textbf{Right:} Prototypical networks transfer features rather than a learning algorithm, and perform poorly if tasks don't have consistent data structure. Training a prototypical network with a fully connected architecture (same as our base model) on a MiniImagenet 10-way classification task with either intact inputs (light purple) or by permuting the pixels before every training and testing task (dark purple). Performance with permuted inputs is greatly reduced (gray line). Our performance is invariant to pixel permutation.
\label{fig:prototypical_network} \label{fig:meta_objective_missmatch}
}
\end{figure}

To illustrate the negative consequences of objective function mismatch in unsupervised learnin algorithms, we train a variational autoencoder on 16x16 CIFAR10. Over the course of training we evaluate classification performance from few shot classification using the learned latent representations.
Training curves can be seen in Figure \ref{fig:meta_objective_missmatch}. Despite continuing to improve the VAE objective throughout training (not shown here), the classification accuracy decreases sharply later in training.

To demonstrate the reduced generalization that results from learning transferable features rather than an update algorithm,  we train a prototypical network \citep{snell2017prototypical} with and without the input shuffling described in Section \ref{sec:sampling_data}. As the prototypical network primarily learns transferrable features, performance is significantly hampered by input shuffling. Results are in Figure \ref{fig:prototypical_network}.

\begin{figure}[h]
\setlength{\belowcaptionskip}{-5pt}
\includegraphics[width=0.80\textwidth]{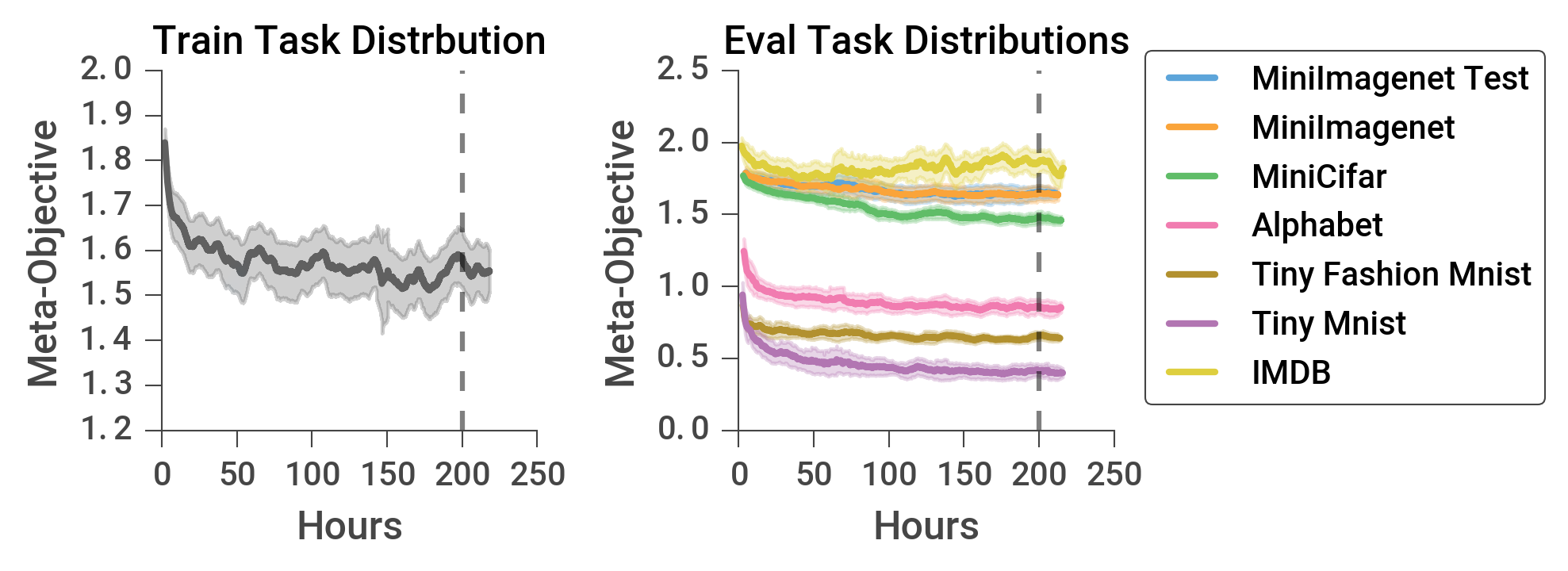}
\centering
\vspace{-5pt}
\caption{Training curves for the training and evaluation task distributions. Our train set consists of MiniImagenet, Alphabet, and MiniCIFAR. Our test sets are Mini Imagenet Test, Tiny Fashion MNIST, Tiny MNIST and IMDB. Error bars denote standard deviation of evaluations with a fixed window of samples evaluated from a single model. Dashed line at 200 hours indicates model used for remaining experiments unless otherwise stated. For a bigger version of this figure, see Appendix \ref{app:meta_optimization}.
\label{fig:theta_training_curve}
}
\end{figure}

\subsection{Meta-Optimization}
While training, we monitor a rolling average of the meta-objective averaged across all datasets, model architectures, and the number of unrolling steps performed. In Figure \ref{fig:theta_training_curve} the training loss is continuing to decrease after 200 hours of training, which suggests that the approximate training techniques still produce effective learning. 
In addition to this global number, we measure performance obtained by rolling out the $\text{UnsupervisedUpdate}$ on various meta-training and meta-testing datasets. We see that on held out image datasets, such as MNIST and Fashion Mnist, the evaluation loss is still decreasing. However, for datasets in a different domain, such as IMDB sentiment prediction \citep{imdb}, we start to see meta-overfitting. For all remaining experimental results, unless otherwise stated, we use meta-parameters, $\theta$, for the $\text{UnsupervisedUpdate}$ resulting from 200 hours of meta-training. 
\vspace{-0.1in}

\subsection{Generalization}
The goal of this work is to learn a general purpose unsupervised representation learning algorithm. As such, this algorithm must be able to generalize across a wide range of scenarios, including tasks that are not sampled i.i.d. from the meta-training distribution. In the following sections, we explore a subset of the factors we seek to generalize over.

\textbf{Generalizing over datasets and domains}

\begin{figure}[h]
\centering
\includegraphics[width=0.51\textwidth]{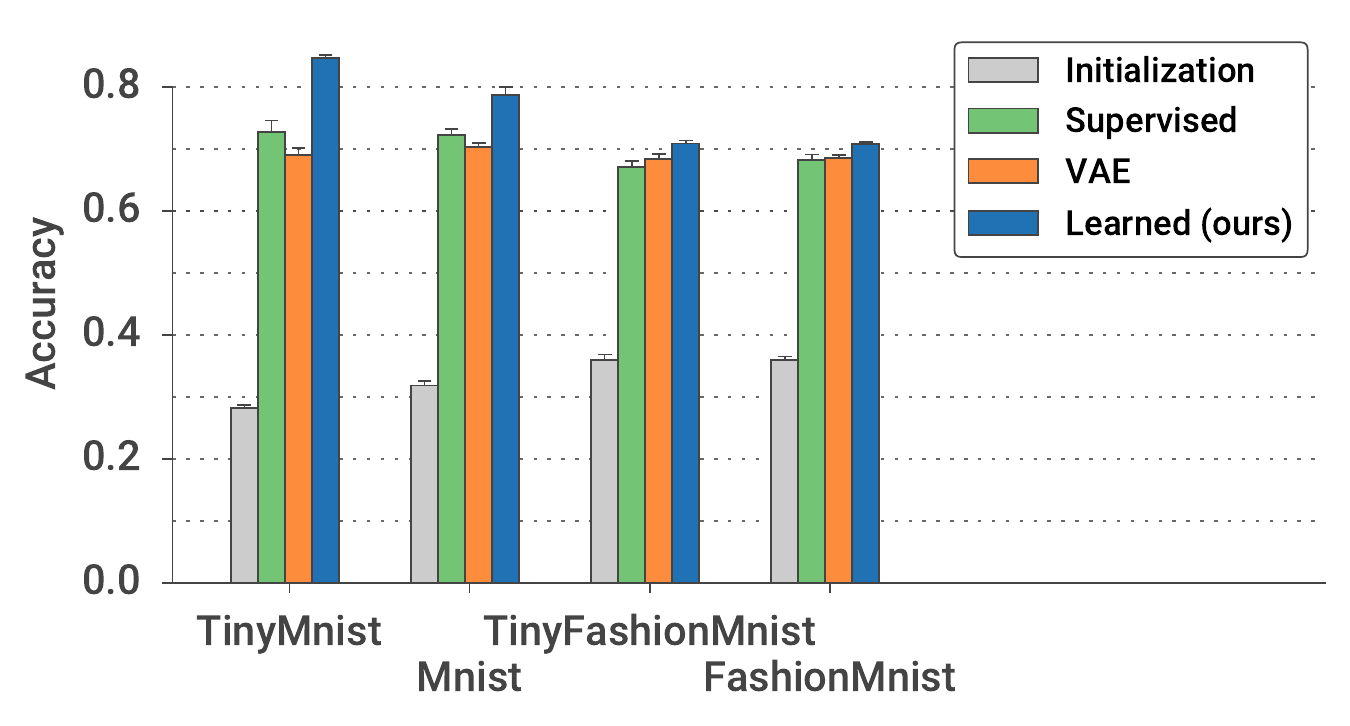}
\setlength{\abovecaptionskip}{-3pt}
\includegraphics[width=0.43\textwidth]{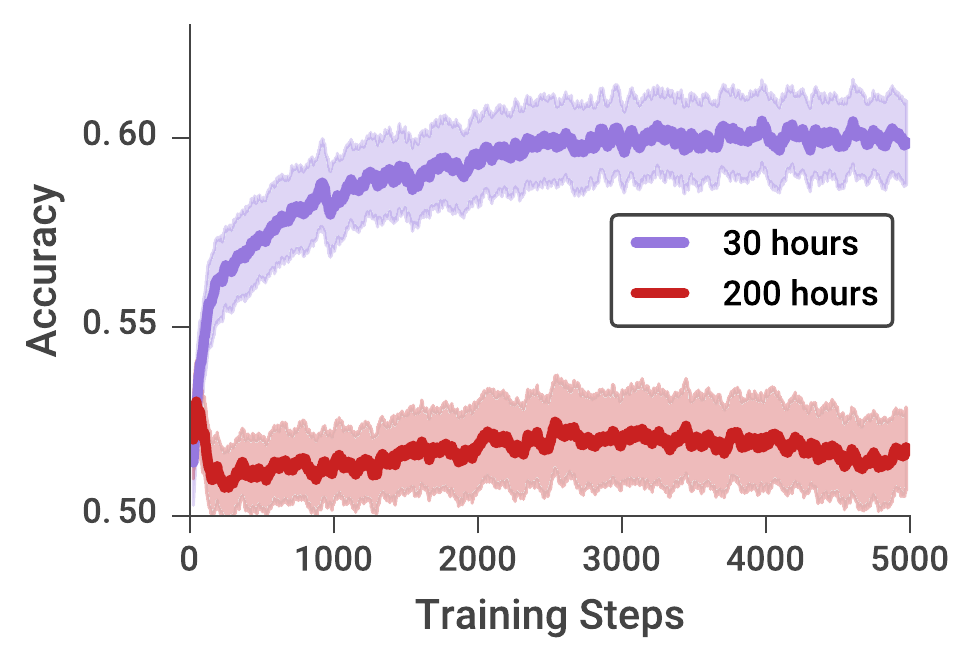}
\caption{
\textbf{Left:} The learned $\text{UnsupervisedUpdate}$ generalizes to unseen datasets.
Our learned update rule produces representations more suitable for few shot classification than those from random initialization or a variational autoecoder and outperforms fully supervised learning on the same labeled examples.
Error bars show standard error.
\label{fig:generalize_dataset}
\textbf{Right:} Early in meta-training (purple), the $\text{UnsupervisedUpdate}$ is able to learn useful features on a 2 way text classification data set, IMDB, despite being meta-trained only from image datasets. Later in meta-training (red) performance drops due to the domain mismatch. We show inner-loop training, consisting of 5k applications of the $\text{UnsupervisedUpdate}$ evaluating the $\text{MetaObjective}$ each iteration. Error bars show standard error across 10 runs.
\label{fig:imdb}
}
\end{figure}

In Figure \ref{fig:generalize_dataset}, we compare performance on few shot classification with 10 examples per class.
We evaluate test performance on holdout datasets of MNIST and Fashion MNIST at 2 resolutions: 14$\times$14 and 28$\times$28 (larger than any dataset experienced in meta-training). 
On the same base model architecture, our learned $\text{UnsupervisedUpdate}$ leads to performance better than a variational autoencoder, supervised learning on the labeled examples, and random initialization with trained readout layer.

To further explore generalization limits, we test our learned optimizer on data from a vastly different domain. We train on a binary text classification dataset: IMDB movie reviews~\citep{imdb}, encoded by computing a bag of words with 1K words. We evaluate using a model 30 hours and 200 hours into meta-training (see Figure \ref{fig:imdb}). Despite being trained exclusively on image datasets, the 30 hour learned optimizer improves upon the random initialization by almost 10\%. When meta-training for longer, however, the learned optimizer ``meta-overfits'' to the image domain resulting in poor performance. This performance is quite low in an absolute sense, for this task.
Nevertheless, we find this result very exciting as we are unaware of any work showing this kind of transfer of learned rules from images to text.

\textbf{Generalizing over network architectures}

\begin{figure}[h]
\centering
\setlength{\belowcaptionskip}{-2pt}
\includegraphics[width=0.52\textwidth,align=c]{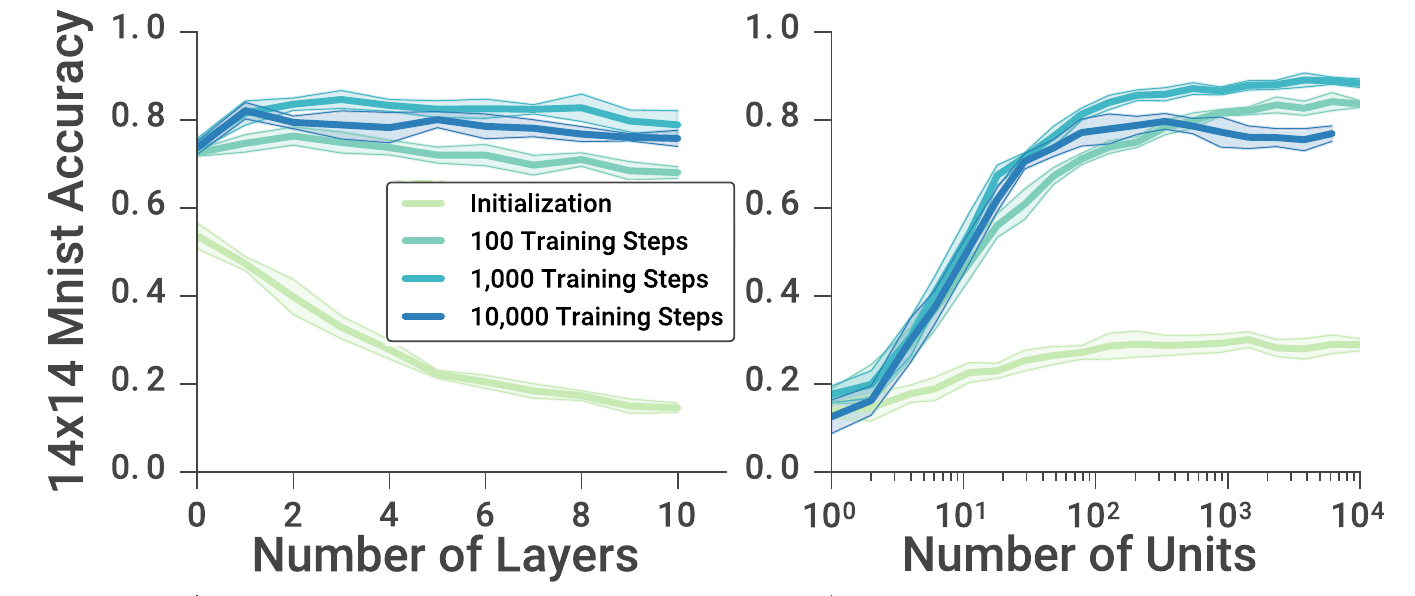}
\vspace{-0.1in}
\includegraphics[width=0.46\textwidth,align=c]{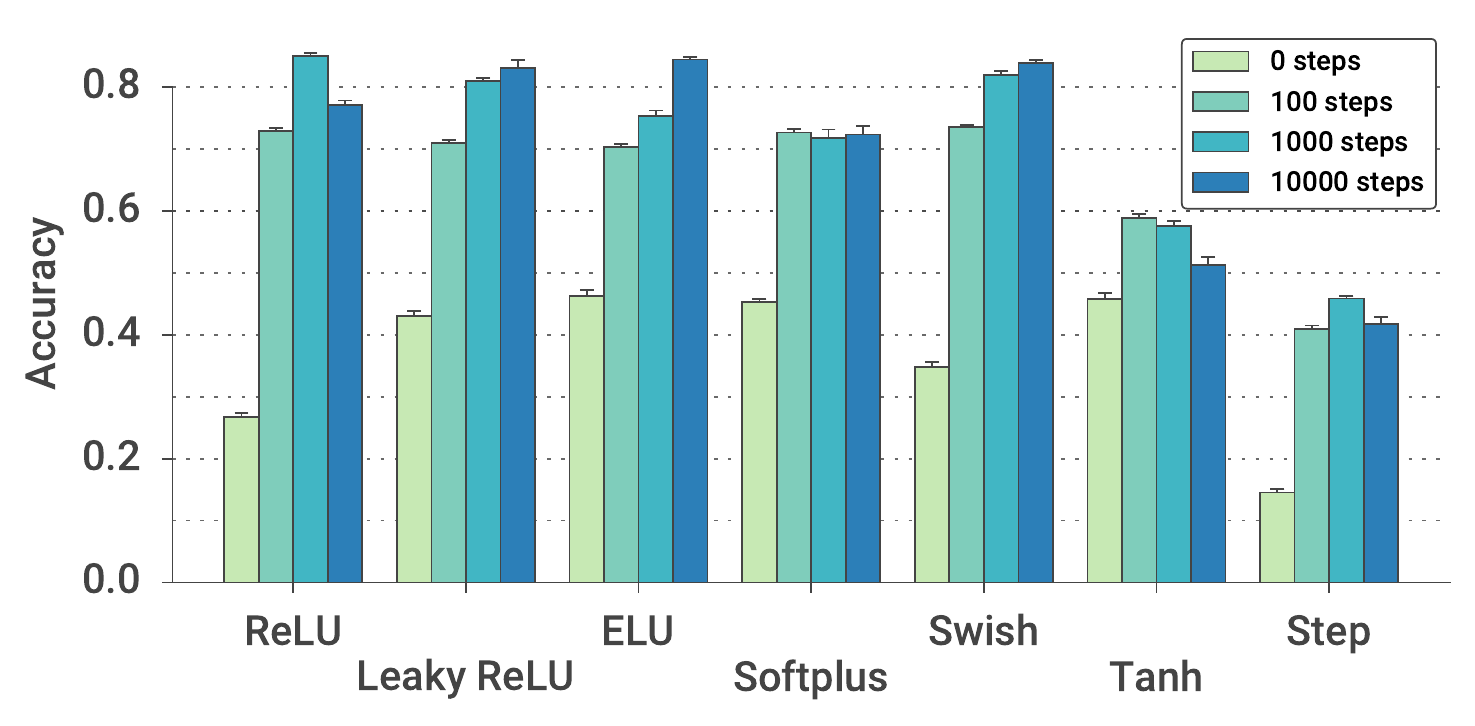}
\caption{\textbf{Left:} The learned $\text{UnsupervisedUpdate}$ is capable of optimizing base models with hidden sizes and depths outside the meta-training regime. As we increase the number of units per layer, the learned model can make use of this additional capacity despite never having experienced it during meta-training. \label{fig:generalize_units_layers}
\textbf{Right:} The learned $\text{UnsupervisedUpdate}$ generalizes across many different activation functions not seen in training. We show accuracy over the course of training on 14x14 MNIST. \label{fig:generalize_activation}
}
\end{figure}

We train models of varying depths and unit counts with our learned optimizer and compare results at different points in training (Figure \ref{fig:generalize_units_layers}). We find that despite only training on networks with 2 to 5 layers and 64 to 512 units per layer, the learned rule generalizes to 11 layers and 10,000 units per layer.

Next we look at generalization over different activation functions. We apply our learned optimizer on base models with a variety of different activation functions. Performance evaluated at different points in training (Figure \ref{fig:generalize_activation}). Despite training only on ReLU activations, our learned optimizer is able to improve on random initializations in all cases. For certain activations, leaky ReLU \citep{maas2013rectifier} and Swish \citep{ramachandran2017searching}, there is little to no decrease in performance. Another interesting case is the step activation function. These activations are traditionally challenging to train as there is no useful gradient signal. Despite this, our learned $\text{UnsupervisedUpdate}$ is capable of optimizing as it does not use base model gradients, and achieves performance double that of random initialization.

\subsection{How it Learns and How it Learns to Learn}
\begin{figure}[h!]
\centering
\setlength{\belowcaptionskip}{-5pt}
\includegraphics[width=0.49\textwidth, align=c]{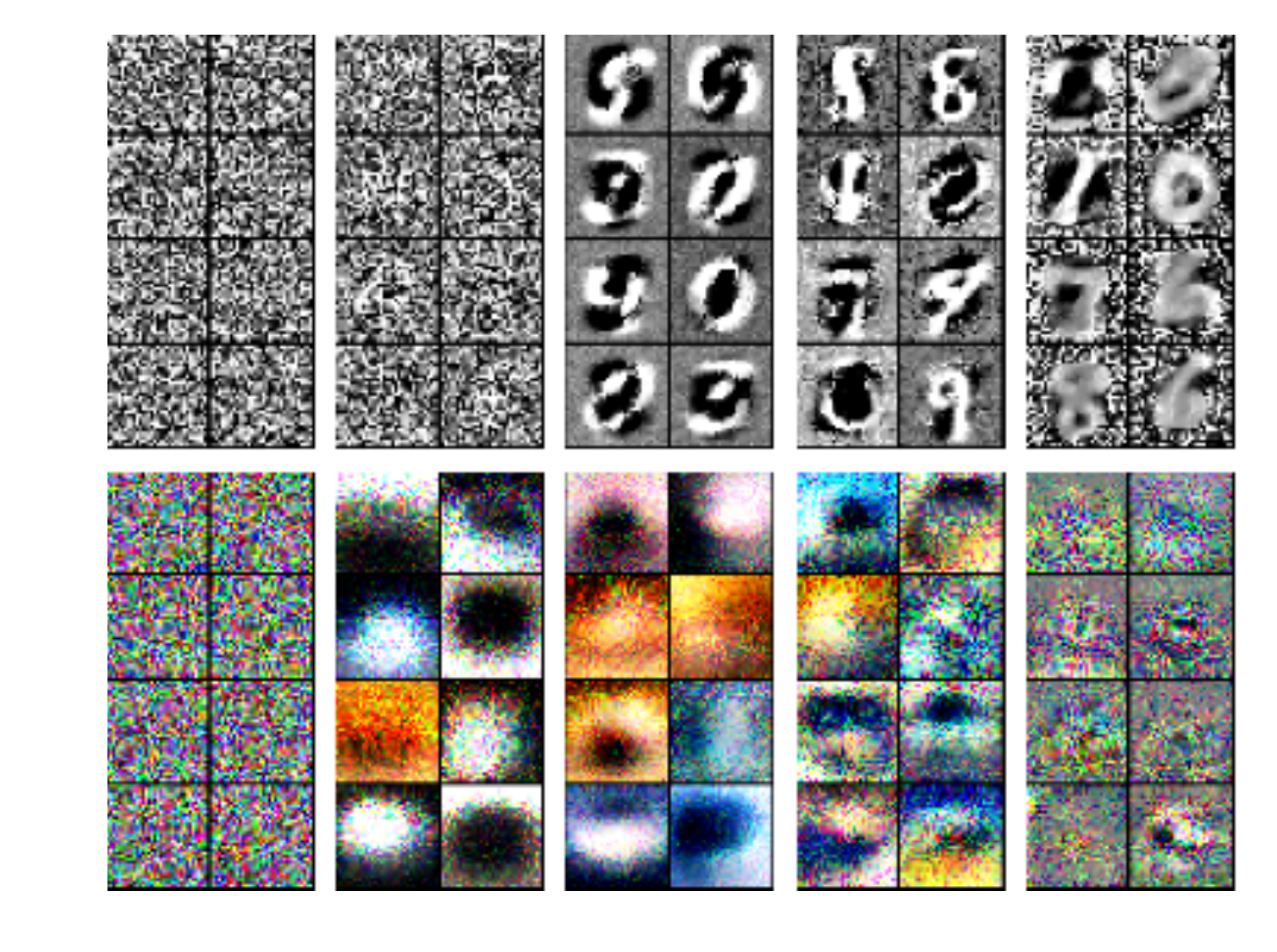}
\includegraphics[width=0.5\textwidth,align=c]{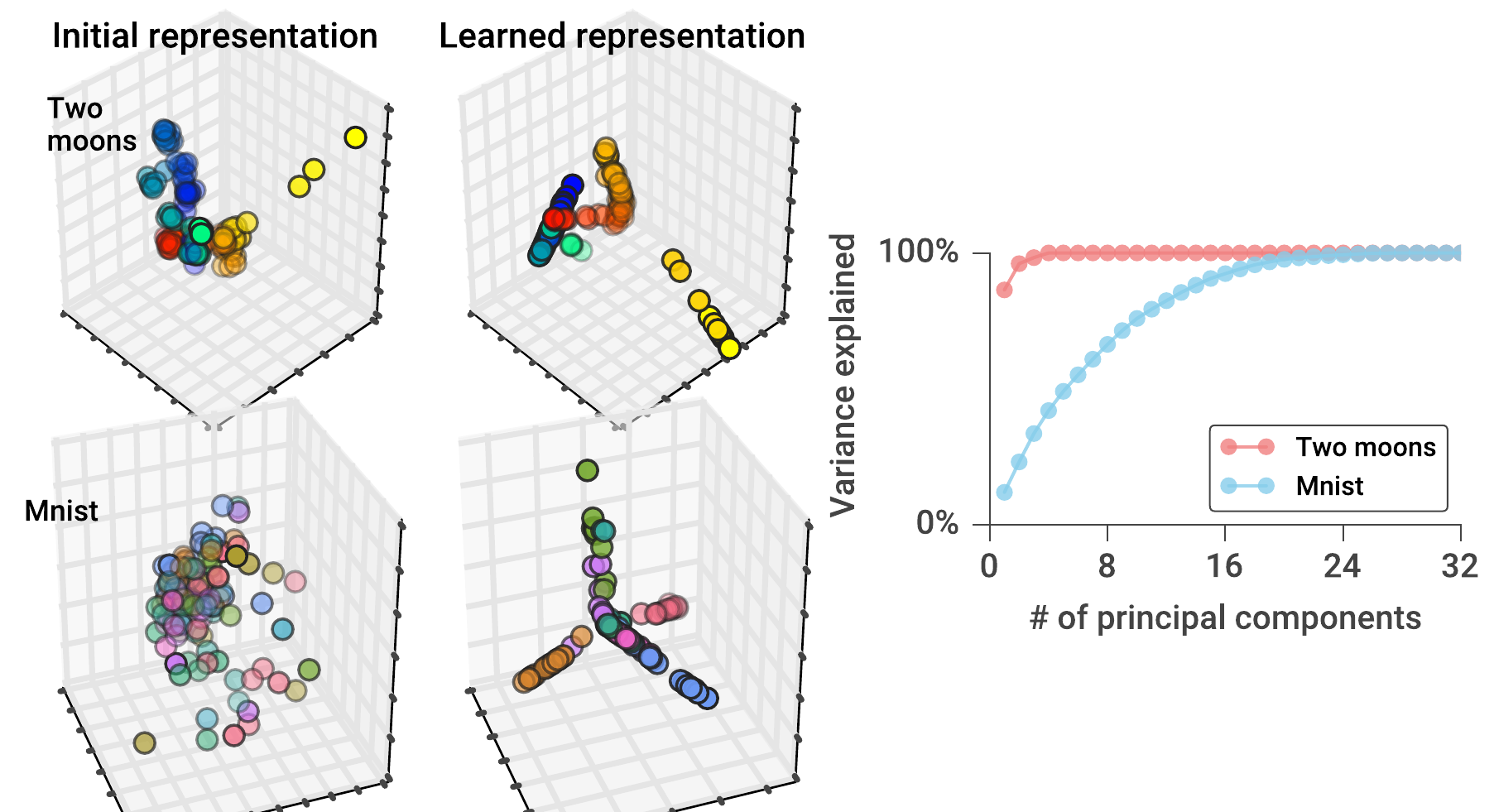}
\vspace{-5pt}
\caption{
\textbf{Left:} From left to right we show first layer base model receptive fields produced by our learned $\text{UnsupervisedUpdate}$ rule over the course of meta-training. Each pane consists of first layer filters extracted from $\phi$ after 10k applications of $\text{UnsupervisedUpdate}$ on MNIST (top) and CIFAR10 (bottom). For MNIST, the optimizer learns image-template-like features. For CIFAR10, low frequency features evolve into higher frequency and more spatially localized features.
For more filters, see Appendix \ref{app:filters}. \label{exp:filters_theta}
\textbf{Center:} Visualization of learned representations before (left) and after (right) training a base model with our learned $\text{UnsupervisedUpdate}$ for two moons (top) and MNIST (bottom). The $\text{UnsupervisedUpdate}$ is capable of manipulating the data manifold, without access to labels, to separate the data classes. Visualization shows a projection of the 32-dimensional representation of the base network onto the top three principal components. \textbf{Right:} Cumulative variance explained using principal components analysis (PCA) on the learned representations. The representation for two moons data (red) is much lower dimensional than MNIST (blue), although both occupy a fraction of the full 32-dimensional space. \label{fig:visualization}
}
\end{figure}

To analyze how our learned optimizer functions, we analyze the first layer filters over the course of meta-training. Despite the permutation invariant nature of our data (enforced by shuffling input image pixels before each unsupervised training run), the base model learns features such as those shown in Figure \ref{exp:filters_theta}, which appear template-like for MNIST, and local-feature-like for CIFAR10. Early in training, there are coarse features, and a lot of noise. As the meta-training progresses, more interesting and local features emerge.

In an effort to understand what our algorithm learns to do, we fed it data from the two moons dataset. We find that despite being a 2D dataset, dissimilar from the image datasets used in meta-training, the learned model is still capable of manipulating and partially separating the data manifold in a purely unsupervised manner (Figure \ref{fig:visualization}). We also find that almost all the variance in the embedding space is dominated by a few dimensions. As a comparison, we do the same analysis on MNIST. In this setting, the explained variance is spread out over more of the principal components. This makes sense as the generative process contains many more latent dimensions -- at least enough to express the 10 digits.

\newpage
\section{Discussion}

In this work we meta-learn an unsupervised representation learning update rule. We show performance that matches or exceeds existing unsupervised learning on held out tasks. Additionally, the update rule can train models of varying widths, depths, and activation functions. More broadly, we demonstrate an application of meta-learning for learning complex optimization tasks where no objective is explicitly defined. Analogously to how increased data and compute have powered supervised learning, we believe this work is a proof of principle that the same can be done with algorithm design--replacing hand designed techniques with architectures designed for learning and learned from data via meta-learning.

\subsubsection*{Acknowledgments}
We would like to thank
Samy Bengio,
David Dohan,
Keren Gu,
Gamaleldin Elsayed,
C. Daniel Freeman, 
Sam Greydanus,
Nando de Freitas,
Ross Goroshin,
Ishaan Gulrajani,
Eric Jang,
Hugo Larochelle,
Jeremy Nixon,
Esteban Real,
Suharsh Sivakumar,
Pavel Sountsov,
Alex Toshev,
George Tucker,
Hoang Trieu Trinh,
Olga Wichrowska,
Lechao Xiao,
Zongheng Yang,
Jiaqi Zhai
and the rest of the Google Brain team for extremely helpful conversations and feedback on this work.

\newpage
\small
\bibliographystyle{unsrtnat}
\bibliography{main}

\begin{thebibliography}{66}
\providecommand{\natexlab}[1]{#1}
\providecommand{\url}[1]{\texttt{#1}}
\expandafter\ifx\csname urlstyle\endcsname\relax
  \providecommand{\doi}[1]{doi: #1}\else
  \providecommand{\doi}{doi: \begingroup \urlstyle{rm}\Url}\fi

\bibitem[Hochreiter et~al.(2001)Hochreiter, Younger, and
  Conwell]{hochreiter2001learning}
Sepp Hochreiter, A~Steven Younger, and Peter~R Conwell.
\newblock Learning to learn using gradient descent.
\newblock In \emph{International Conference on Artificial Neural Networks},
  pages 87--94. Springer, 2001.

\bibitem[Schmidhuber(1995)]{schmidhuber1995learning}
Juergen Schmidhuber.
\newblock On learning how to learn learning strategies.
\newblock 1995.

\bibitem[Hinton and Salakhutdinov(2006)]{hinton2006reducing}
Geoffrey~E Hinton and Ruslan~R Salakhutdinov.
\newblock Reducing the dimensionality of data with neural networks.
\newblock \emph{science}, 313\penalty0 (5786):\penalty0 504--507, 2006.

\bibitem[Vincent et~al.(2008)Vincent, Larochelle, Bengio, and
  Manzagol]{vincent2008extracting}
Pascal Vincent, Hugo Larochelle, Yoshua Bengio, and Pierre-Antoine Manzagol.
\newblock Extracting and composing robust features with denoising autoencoders.
\newblock In \emph{Proceedings of the 25th international conference on Machine
  learning}, pages 1096--1103. ACM, 2008.

\bibitem[Vincent et~al.(2010)Vincent, Larochelle, Lajoie, Bengio, and
  Manzagol]{vincent2010stacked}
Pascal Vincent, Hugo Larochelle, Isabelle Lajoie, Yoshua Bengio, and
  Pierre-Antoine Manzagol.
\newblock Stacked denoising autoencoders: Learning useful representations in a
  deep network with a local denoising criterion.
\newblock \emph{Journal of Machine Learning Research}, 11\penalty0
  (Dec):\penalty0 3371--3408, 2010.

\bibitem[Kingma and Welling(2013)]{kingma2013auto}
Diederik~P Kingma and Max Welling.
\newblock Auto-encoding variational bayes.
\newblock \emph{arXiv preprint arXiv:1312.6114}, 2013.

\bibitem[Le et~al.(2011)Le, Ranzato, Monga, Devin, Chen, Corrado, Dean, and
  Ng]{le2011building}
Quoc~V Le, Marc'Aurelio Ranzato, Rajat Monga, Matthieu Devin, Kai Chen, Greg~S
  Corrado, Jeff Dean, and Andrew~Y Ng.
\newblock Building high-level features using large scale unsupervised learning.
\newblock \emph{arXiv preprint arXiv:1112.6209}, 2011.

\bibitem[Goodfellow et~al.(2014)Goodfellow, Pouget-Abadie, Mirza, Xu,
  Warde-Farley, Ozair, Courville, and Bengio]{goodfellow2014generative}
Ian Goodfellow, Jean Pouget-Abadie, Mehdi Mirza, Bing Xu, David Warde-Farley,
  Sherjil Ozair, Aaron Courville, and Yoshua Bengio.
\newblock Generative adversarial nets.
\newblock In \emph{Advances in neural information processing systems}, pages
  2672--2680, 2014.

\bibitem[Radford et~al.(2015)Radford, Metz, and
  Chintala]{radford2015unsupervised}
Alec Radford, Luke Metz, and Soumith Chintala.
\newblock Unsupervised representation learning with deep convolutional
  generative adversarial networks.
\newblock \emph{arXiv preprint arXiv:1511.06434}, 2015.

\bibitem[Donahue et~al.(2016)Donahue, Kr{\"a}henb{\"u}hl, and
  Darrell]{donahue2016adversarial}
Jeff Donahue, Philipp Kr{\"a}henb{\"u}hl, and Trevor Darrell.
\newblock Adversarial feature learning.
\newblock \emph{arXiv preprint arXiv:1605.09782}, 2016.

\bibitem[Dumoulin et~al.(2016)Dumoulin, Belghazi, Poole, Lamb, Arjovsky,
  Mastropietro, and Courville]{dumoulin2016adversarially}
Vincent Dumoulin, Ishmael Belghazi, Ben Poole, Alex Lamb, Martin Arjovsky,
  Olivier Mastropietro, and Aaron Courville.
\newblock Adversarially learned inference.
\newblock \emph{arXiv preprint arXiv:1606.00704}, 2016.

\bibitem[Noroozi and Favaro(2016)]{noroozi2016unsupervised}
Mehdi Noroozi and Paolo Favaro.
\newblock Unsupervised learning of visual representations by solving jigsaw
  puzzles.
\newblock In \emph{European Conference on Computer Vision}, pages 69--84.
  Springer, 2016.

\bibitem[Misra et~al.(2016)Misra, Zitnick, and Hebert]{misra2016shuffle}
Ishan Misra, C~Lawrence Zitnick, and Martial Hebert.
\newblock Shuffle and learn: unsupervised learning using temporal order
  verification.
\newblock In \emph{European Conference on Computer Vision}, pages 527--544.
  Springer, 2016.

\bibitem[Sermanet et~al.(2017)Sermanet, Lynch, Hsu, and
  Levine]{sermanet2017time}
Pierre Sermanet, Corey Lynch, Jasmine Hsu, and Sergey Levine.
\newblock Time-contrastive networks: Self-supervised learning from multi-view
  observation.
\newblock \emph{arXiv preprint arXiv:1704.06888}, 2017.

\bibitem[Coates and Ng(2012)]{coates2012learning}
Adam Coates and Andrew~Y Ng.
\newblock Learning feature representations with k-means.
\newblock In \emph{Neural networks: Tricks of the trade}, pages 561--580.
  Springer, 2012.

\bibitem[Xie et~al.(2016)Xie, Girshick, and Farhadi]{xie2016unsupervised}
Junyuan Xie, Ross Girshick, and Ali Farhadi.
\newblock Unsupervised deep embedding for clustering analysis.
\newblock In \emph{International conference on machine learning}, pages
  478--487, 2016.

\bibitem[Bojanowski and Joulin(2017)]{bojanowski2017unsupervised}
Piotr Bojanowski and Armand Joulin.
\newblock Unsupervised learning by predicting noise.
\newblock \emph{arXiv preprint arXiv:1704.05310}, 2017.

\bibitem[Schmidhuber(1992)]{schmidhuber1992learning}
J{\"u}rgen Schmidhuber.
\newblock Learning factorial codes by predictability minimization.
\newblock \emph{Neural Computation}, 4\penalty0 (6):\penalty0 863--879, 1992.

\bibitem[Hochreiter and Schmidhuber(1999)]{hochreiter1999feature}
Sepp Hochreiter and J{\"u}rgen Schmidhuber.
\newblock Feature extraction through lococode.
\newblock \emph{Neural Computation}, 11\penalty0 (3):\penalty0 679--714, 1999.

\bibitem[Olshausen and Field(1997)]{olshausen1997sparse}
Bruno~A Olshausen and David~J Field.
\newblock Sparse coding with an overcomplete basis set: A strategy employed by
  v1?
\newblock \emph{Vision research}, 37\penalty0 (23):\penalty0 3311--3325, 1997.

\bibitem[Schmidhuber(1987)]{schmidhuber1987evolutionary}
J{\"u}rgen Schmidhuber.
\newblock \emph{Evolutionary principles in self-referential learning, or on
  learning how to learn: the meta-meta-... hook}.
\newblock PhD thesis, Technische Universit{\"a}t M{\"u}nchen, 1987.

\bibitem[Bengio et~al.(1990)Bengio, Bengio, and Cloutier]{bengio1990learning}
Yoshua Bengio, Samy Bengio, and Jocelyn Cloutier.
\newblock \emph{Learning a synaptic learning rule}.
\newblock Universit{\'e} de Montr{\'e}al, D{\'e}partement d'informatique et de
  recherche op{\'e}rationnelle, 1990.

\bibitem[Bengio et~al.(1992)Bengio, Bengio, Cloutier, and
  Gecsei]{bengio1992optimization}
Samy Bengio, Yoshua Bengio, Jocelyn Cloutier, and Jan Gecsei.
\newblock On the optimization of a synaptic learning rule.
\newblock In \emph{Preprints Conf. Optimality in Artificial and Biological
  Neural Networks}, pages 6--8. Univ. of Texas, 1992.

\bibitem[Runarsson and Jonsson(2000)]{runarsson2000evolution}
Thomas~Philip Runarsson and Magnus~Thor Jonsson.
\newblock Evolution and design of distributed learning rules.
\newblock In \emph{Combinations of Evolutionary Computation and Neural
  Networks, 2000 IEEE Symposium on}, pages 59--63. IEEE, 2000.

\bibitem[Vinyals et~al.(2016)Vinyals, Blundell, Lillicrap, kavukcuoglu, and
  Wierstra]{Vinyals2016}
Oriol Vinyals, Charles Blundell, Tim Lillicrap, koray kavukcuoglu, and Daan
  Wierstra.
\newblock Matching networks for one shot learning.
\newblock In D.~D. Lee, M.~Sugiyama, U.~V. Luxburg, I.~Guyon, and R.~Garnett,
  editors, \emph{Advances in Neural Information Processing Systems 29}, pages
  3630--3638. Curran Associates, Inc., 2016.
\newblock URL
  \url{http://papers.nips.cc/paper/6385-matching-networks-for-one-shot-learning.pdf}.

\bibitem[Ravi and Larochelle(2016)]{ravi2016optimization}
Sachin Ravi and Hugo Larochelle.
\newblock Optimization as a model for few-shot learning.
\newblock \emph{International Conference on Learning Representations}, 2016.

\bibitem[Mishra et~al.(2017)Mishra, Rohaninejad, Chen, and
  Abbeel]{mishra2017meta}
Nikhil Mishra, Mostafa Rohaninejad, Xi~Chen, and Pieter Abbeel.
\newblock Meta-learning with temporal convolutions.
\newblock \emph{arXiv preprint arXiv:1707.03141}, 2017.

\bibitem[Finn et~al.(2017)Finn, Abbeel, and Levine]{finn17}
Chelsea Finn, Pieter Abbeel, and Sergey Levine.
\newblock Model-agnostic meta-learning for fast adaptation of deep networks.
\newblock In Doina Precup and Yee~Whye Teh, editors, \emph{Proceedings of the
  34th International Conference on Machine Learning}, volume~70 of
  \emph{Proceedings of Machine Learning Research}, pages 1126--1135,
  International Convention Centre, Sydney, Australia, 06--11 Aug 2017. PMLR.
\newblock URL \url{http://proceedings.mlr.press/v70/finn17a.html}.

\bibitem[Snell et~al.(2017)Snell, Swersky, and Zemel]{snell2017prototypical}
Jake Snell, Kevin Swersky, and Richard Zemel.
\newblock Prototypical networks for few-shot learning.
\newblock In \emph{Advances in Neural Information Processing Systems}, pages
  4080--4090, 2017.

\bibitem[Ren et~al.(2018)Ren, Triantafillou, Ravi, Snell, Swersky, Tenenbaum,
  Larochelle, and Zemel]{ren2018meta}
Mengye Ren, Eleni Triantafillou, Sachin Ravi, Jake Snell, Kevin Swersky,
  Joshua~B Tenenbaum, Hugo Larochelle, and Richard~S Zemel.
\newblock Meta-learning for semi-supervised few-shot classification.
\newblock \emph{arXiv preprint arXiv:1803.00676}, 2018.

\bibitem[Hsu et~al.(2018)Hsu, Levine, and Finn]{hsu2018unsupervised}
Kyle Hsu, Sergey Levine, and Chelsea Finn.
\newblock Unsupervised learning via meta-learning.
\newblock \emph{arXiv preprint arXiv:1810.02334}, 2018.

\bibitem[Garg(2018)]{garg2017supervising}
Vikas Garg.
\newblock Supervising unsupervised learning.
\newblock In \emph{Advances in Neural Information Processing Systems}, pages
  4996--5006, 2018.

\bibitem[Jones(2001)]{jones2001taxonomy}
Donald~R Jones.
\newblock A taxonomy of global optimization methods based on response surfaces.
\newblock \emph{Journal of global optimization}, 21\penalty0 (4):\penalty0
  345--383, 2001.

\bibitem[Snoek et~al.(2012)Snoek, Larochelle, and Adams]{snoek2012practical}
Jasper Snoek, Hugo Larochelle, and Ryan~P Adams.
\newblock Practical bayesian optimization of machine learning algorithms.
\newblock In \emph{Advances in neural information processing systems}, pages
  2951--2959, 2012.

\bibitem[Bergstra et~al.(2011)Bergstra, Bardenet, Bengio, and
  K{\'e}gl]{bergstra2011algorithms}
James~S Bergstra, R{\'e}mi Bardenet, Yoshua Bengio, and Bal{\'a}zs K{\'e}gl.
\newblock Algorithms for hyper-parameter optimization.
\newblock In \emph{Advances in neural information processing systems}, pages
  2546--2554, 2011.

\bibitem[Bergstra and Bengio(2012)]{bergstra2012random}
James Bergstra and Yoshua Bengio.
\newblock Random search for hyper-parameter optimization.
\newblock \emph{Journal of Machine Learning Research}, 13\penalty0
  (Feb):\penalty0 281--305, 2012.

\bibitem[Stanley and Miikkulainen(2002)]{stanley2002evolving}
Kenneth~O Stanley and Risto Miikkulainen.
\newblock Evolving neural networks through augmenting topologies.
\newblock \emph{Evolutionary computation}, 10\penalty0 (2):\penalty0 99--127,
  2002.

\bibitem[Zoph and Le(2017)]{Zoph2017}
Barret Zoph and Quoc~V. Le.
\newblock Neural architecture search with reinforcement learning.
\newblock \emph{International Conference on Learning Representations}, 2017.
\newblock URL \url{https://arxiv.org/abs/1611.01578}.

\bibitem[Baker et~al.(2017)Baker, Gupta, Naik, and Raskar]{baker2016designing}
Bowen Baker, Otkrist Gupta, Nikhil Naik, and Ramesh Raskar.
\newblock Designing neural network architectures using reinforcement learning.
\newblock \emph{International Conference on Learning Representations}, 2017.

\bibitem[Zoph et~al.(2018)Zoph, Vasudevan, Shlens, and Le]{zoph2017learning}
Barret Zoph, Vijay Vasudevan, Jonathon Shlens, and Quoc~V Le.
\newblock Learning transferable architectures for scalable image recognition.
\newblock \emph{Proceedings of the IEEE conference on computer vision and
  pattern recognition}, 2018.

\bibitem[Real et~al.(2017)Real, Moore, Selle, Saxena, Suematsu, Le, and
  Kurakin]{real2017large}
Esteban Real, Sherry Moore, Andrew Selle, Saurabh Saxena, Yutaka~Leon Suematsu,
  Quoc Le, and Alex Kurakin.
\newblock Large-scale evolution of image classifiers.
\newblock \emph{arXiv preprint arXiv:1703.01041}, 2017.

\bibitem[Maclaurin et~al.(2015)Maclaurin, Duvenaud, and
  Adams]{maclaurin2015gradient}
Dougal Maclaurin, David Duvenaud, and Ryan Adams.
\newblock Gradient-based hyperparameter optimization through reversible
  learning.
\newblock In \emph{International Conference on Machine Learning}, pages
  2113--2122, 2015.

\bibitem[Andrychowicz et~al.(2016)Andrychowicz, Denil, Gomez, Hoffman, Pfau,
  Schaul, and de~Freitas]{andrychowicz2016learning}
Marcin Andrychowicz, Misha Denil, Sergio Gomez, Matthew~W Hoffman, David Pfau,
  Tom Schaul, and Nando de~Freitas.
\newblock Learning to learn by gradient descent by gradient descent.
\newblock In \emph{Advances in Neural Information Processing Systems}, pages
  3981--3989, 2016.

\bibitem[Chen et~al.(2016)Chen, Hoffman, Colmenarejo, Denil, Lillicrap,
  Botvinick, and de~Freitas]{chen2016learning}
Yutian Chen, Matthew~W Hoffman, Sergio~G{\'o}mez Colmenarejo, Misha Denil,
  Timothy~P Lillicrap, Matt Botvinick, and Nando de~Freitas.
\newblock Learning to learn without gradient descent by gradient descent.
\newblock \emph{arXiv preprint arXiv:1611.03824}, 2016.

\bibitem[Li and Malik(2017)]{li2016learning}
Ke~Li and Jitendra Malik.
\newblock Learning to optimize.
\newblock \emph{International Conference on Learning Representations}, 2017.

\bibitem[Wichrowska et~al.(2017)Wichrowska, Maheswaranathan, Hoffman,
  Colmenarejo, Denil, de~Freitas, and Sohl-Dickstein]{wichrowska2017learned}
Olga Wichrowska, Niru Maheswaranathan, Matthew~W Hoffman, Sergio~Gomez
  Colmenarejo, Misha Denil, Nando de~Freitas, and Jascha Sohl-Dickstein.
\newblock Learned optimizers that scale and generalize.
\newblock \emph{International Conference on Machine Learning}, 2017.

\bibitem[Bello et~al.(2017)Bello, Zoph, Vasudevan, and Le]{Bello17}
Irwan Bello, Barret Zoph, Vijay Vasudevan, and Quoc Le.
\newblock Neural optimizer search with reinforcement learning.
\newblock 2017.
\newblock URL \url{https://arxiv.org/pdf/1709.07417.pdf}.

\bibitem[Houthooft et~al.(2018)Houthooft, Chen, Isola, Stadie, Wolski, Ho, and
  Abbeel]{houthooft2018evolved}
Rein Houthooft, Richard~Y Chen, Phillip Isola, Bradly~C Stadie, Filip Wolski,
  Jonathan Ho, and Pieter Abbeel.
\newblock Evolved policy gradients.
\newblock \emph{arXiv preprint arXiv:1802.04821}, 2018.

\bibitem[Ioffe and Szegedy(2015)]{Ioffe15}
Sergey Ioffe and Christian Szegedy.
\newblock Batch normalization: Accelerating deep network training by reducing
  internal covariate shift.
\newblock In Francis Bach and David Blei, editors, \emph{Proceedings of the
  32nd International Conference on Machine Learning}, volume~37 of
  \emph{Proceedings of Machine Learning Research}, pages 448--456, Lille,
  France, 07--09 Jul 2015. PMLR.
\newblock URL \url{http://proceedings.mlr.press/v37/ioffe15.html}.

\bibitem[Whittington and Bogacz(2017)]{whittington2017approximation}
James~CR Whittington and Rafal Bogacz.
\newblock An approximation of the error backpropagation algorithm in a
  predictive coding network with local hebbian synaptic plasticity.
\newblock \emph{Neural computation}, 29\penalty0 (5):\penalty0 1229--1262,
  2017.

\bibitem[Lillicrap et~al.(2016)Lillicrap, Cownden, Tweed, and
  Akerman]{lillicrap2016random}
Timothy~P Lillicrap, Daniel Cownden, Douglas~B Tweed, and Colin~J Akerman.
\newblock Random synaptic feedback weights support error backpropagation for
  deep learning.
\newblock \emph{Nature communications}, 7:\penalty0 13276, 2016.

\bibitem[Pascanu et~al.(2013)Pascanu, Mikolov, and
  Bengio]{pascanu2013difficulty}
Razvan Pascanu, Tomas Mikolov, and Yoshua Bengio.
\newblock On the difficulty of training recurrent neural networks.
\newblock In \emph{International Conference on Machine Learning}, pages
  1310--1318, 2013.

\bibitem[Shaban et~al.(2018)Shaban, Cheng, Hatch, and
  Boots]{shaban2018truncated}
Amirreza Shaban, Ching-An Cheng, Nathan Hatch, and Byron Boots.
\newblock Truncated back-propagation for bilevel optimization.
\newblock \emph{arXiv preprint arXiv:1810.10667}, 2018.

\bibitem[Krizhevsky and Hinton(2009)]{krizhevsky2009learning}
Alex Krizhevsky and Geoffrey Hinton.
\newblock Learning multiple layers of features from tiny images.
\newblock 2009.

\bibitem[Russakovsky et~al.(2015)Russakovsky, Deng, Su, Krause, Satheesh, Ma,
  Huang, Karpathy, Khosla, Bernstein, Berg, and Fei-Fei]{ILSVRC15}
Olga Russakovsky, Jia Deng, Hao Su, Jonathan Krause, Sanjeev Satheesh, Sean Ma,
  Zhiheng Huang, Andrej Karpathy, Aditya Khosla, Michael Bernstein,
  Alexander~C. Berg, and Li~Fei-Fei.
\newblock {ImageNet Large Scale Visual Recognition Challenge}.
\newblock \emph{International Journal of Computer Vision (IJCV)}, 115\penalty0
  (3):\penalty0 211--252, 2015.
\newblock \doi{10.1007/s11263-015-0816-y}.

\bibitem[LeCun et~al.(1998)LeCun, Bottou, Bengio, and
  Haffner]{lecun1998gradient}
Yann LeCun, L{\'e}on Bottou, Yoshua Bengio, and Patrick Haffner.
\newblock Gradient-based learning applied to document recognition.
\newblock \emph{Proceedings of the IEEE}, 86\penalty0 (11):\penalty0
  2278--2324, 1998.

\bibitem[Xiao et~al.(2017)Xiao, Rasul, and Vollgraf]{xiao2017/online}
Han Xiao, Kashif Rasul, and Roland Vollgraf.
\newblock Fashion-mnist: a novel image dataset for benchmarking machine
  learning algorithms, 2017.

\bibitem[Maas et~al.(2011)Maas, Daly, Pham, Huang, Ng, and Potts]{imdb}
Andrew~L. Maas, Raymond~E. Daly, Peter~T. Pham, Dan Huang, Andrew~Y. Ng, and
  Christopher Potts.
\newblock Learning word vectors for sentiment analysis.
\newblock In \emph{Proceedings of the 49th Annual Meeting of the Association
  for Computational Linguistics: Human Language Technologies}, pages 142--150,
  Portland, Oregon, USA, June 2011. Association for Computational Linguistics.
\newblock URL \url{http://www.aclweb.org/anthology/P11-1015}.

\bibitem[Abadi et~al.(2016)Abadi, Barham, Chen, Chen, Davis, Dean, Devin,
  Ghemawat, Irving, Isard, et~al.]{abadi2016tensorflow}
Mart{\'\i}n Abadi, Paul Barham, Jianmin Chen, Zhifeng Chen, Andy Davis, Jeffrey
  Dean, Matthieu Devin, Sanjay Ghemawat, Geoffrey Irving, Michael Isard, et~al.
\newblock Tensorflow: A system for large-scale machine learning.
\newblock In \emph{OSDI}, volume~16, pages 265--283, 2016.

\bibitem[Maas et~al.(2013)Maas, Hannun, and Ng]{maas2013rectifier}
Andrew~L Maas, Awni~Y Hannun, and Andrew~Y Ng.
\newblock Rectifier nonlinearities improve neural network acoustic models.
\newblock In \emph{Proc. icml}, volume~30, page~3, 2013.

\bibitem[Ramachandran et~al.(2017)Ramachandran, Zoph, and
  Le]{ramachandran2017searching}
Prajit Ramachandran, Barret Zoph, and Quoc Le.
\newblock Searching for activation functions.
\newblock 2017.

\bibitem[Pascanu et~al.(2012)Pascanu, Mikolov, and
  Bengio]{pascanu2012understanding}
Razvan Pascanu, Tomas Mikolov, and Yoshua Bengio.
\newblock Understanding the exploding gradient problem.
\newblock \emph{CoRR, abs/1211.5063}, 2012.

\bibitem[Kingma and Ba(2014)]{kingma2014adam}
Diederik~P Kingma and Jimmy Ba.
\newblock Adam: A method for stochastic optimization.
\newblock \emph{arXiv preprint arXiv:1412.6980}, 2014.

\bibitem[Mnih et~al.(2013)Mnih, Kavukcuoglu, Silver, Graves, Antonoglou,
  Wierstra, and Riedmiller]{mnih2013playing}
Volodymyr Mnih, Koray Kavukcuoglu, David Silver, Alex Graves, Ioannis
  Antonoglou, Daan Wierstra, and Martin Riedmiller.
\newblock Playing atari with deep reinforcement learning.
\newblock \emph{arXiv preprint arXiv:1312.5602}, 2013.

\bibitem[Pearlmutter(1996)]{pearlmutter1996investigation}
Barak Pearlmutter.
\newblock \emph{An investigation of the gradient descent process in neural
  networks}.
\newblock PhD thesis, Carnegie Mellon University Pittsburgh, PA, 1996.

\bibitem[Schoenholz et~al.(2016)Schoenholz, Gilmer, Ganguli, and
  Sohl-Dickstein]{schoenholz2016deep}
Samuel~S Schoenholz, Justin Gilmer, Surya Ganguli, and Jascha Sohl-Dickstein.
\newblock Deep information propagation.
\newblock \emph{arXiv preprint arXiv:1611.01232}, 2016.

\end{thebibliography}

\newpage

\normalsize
\onecolumn
\appendix
\setcounter{equation}{0}
\setcounter{figure}{0}
\setcounter{table}{0}
\makeatletter
\renewcommand{\theequation}{App.\arabic{equation}}
\renewcommand{\thefigure}{App.\arabic{figure}}

\section{More Detailed System Diagram}
\label{app:detail_schematic}
\begin{figure}[h!]
\centering
\includegraphics[width=0.50\textwidth]{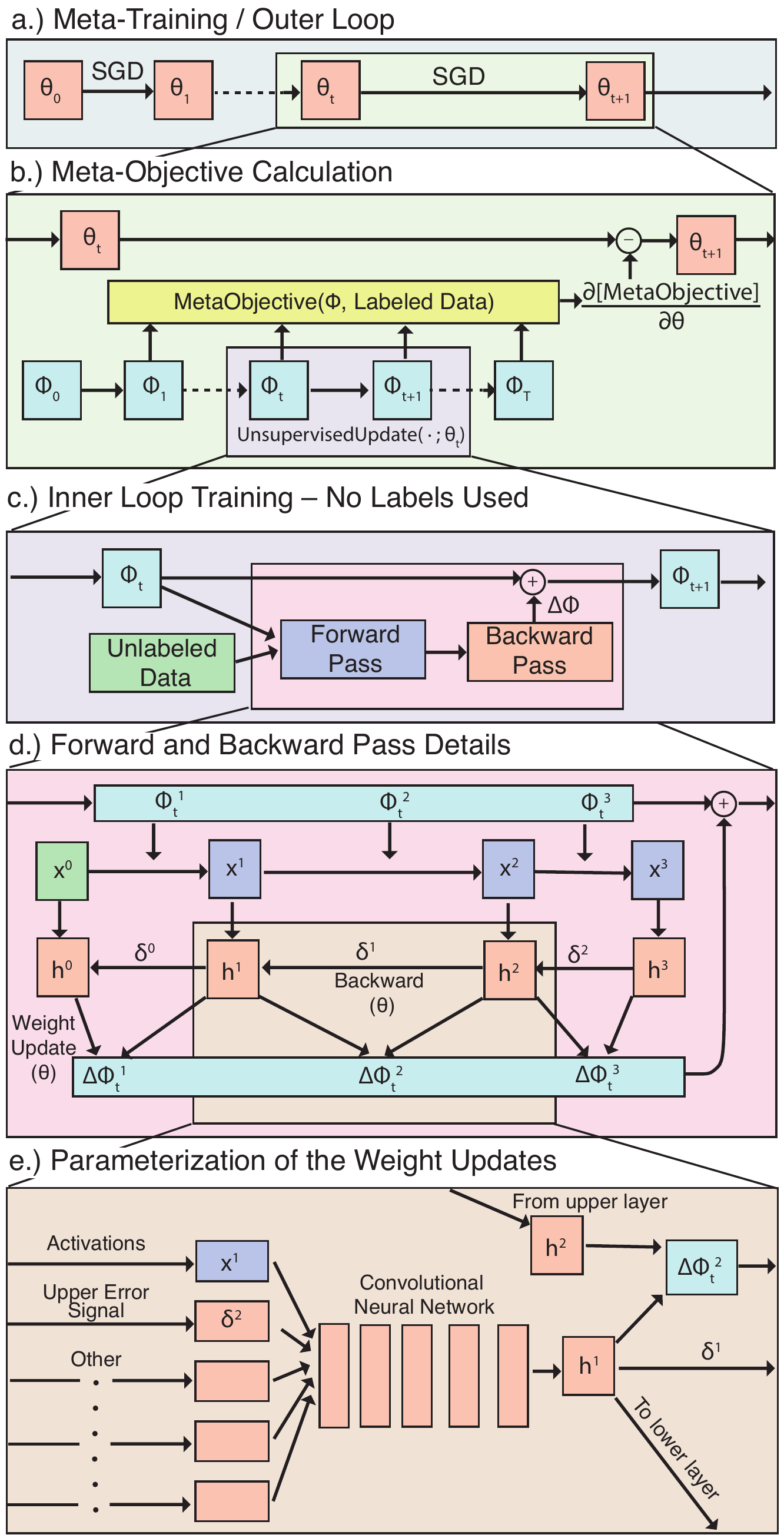}
\caption{Schematic for meta-learning an unsupervised learning algorithm. We show the hierarchical nature of both the meta-training procedure and update rule.
\textbf{a)} Meta-training, where the meta-parameters, $\theta$, are updated via our meta-optimizer (SGD). 
\textbf{b)} The gradients of the $\text{MetaObjective}$ with respect to $\theta$ are computed by backpropagation through the unrolled application of the $\text{UnsupervisedUpdate}$.
\textbf{c)} $\text{UnsupervisedUpdate}$ updates the base model parameters ($\phi$) using a minibatch of unlabeled data. 
\textbf{d)} Each application of $\text{UnsupervisedUpdate}$ involves computing a forward and ``backward'' pass through the base model. 
The base model itself is a fully connected network producing hidden states $x^l$ for each layer $l$. 
The ``backward'' pass through the base model uses an error signal from the layer above, $\delta$, which is generated by a meta-learned function. 
\textbf{e.)} The weight updates $\Delta \phi$ are computed using a convolutional network, using $\delta$ and $x$ from the pre- and post-synaptic neurons, along with several other terms discussed in the text.}
\end{figure}

\section{Stabilizing gradient based meta-learning training}\label{app stability}

Training and computing derivatives through recurrent computation of this form is notoriously difficult \cite{pascanu2013difficulty}.
Training parameters of recurrent systems in general can lead to chaos. We used the usual techniques such as gradient clipping \citep{pascanu2012understanding}, small learning rates, and adaptive learning rate methods (in our case Adam \citep{kingma2014adam}), but in practice this was not enough to train most $\text{UnsupervisedUpdate}$ architectures. In this section we address other techniques needed for stable convergence.

When training with truncated backprop the problem shifts from pure optimization to something more like optimizing on a Markov Decision Process where the state space is the base-model weights, $\phi$, and the `policy' is the learned optimizer. While traversing these states, the policy is constantly meta-optimized and changing, thus changing the distribution of states the optimizer reaches. 
This type of non-i.i.d training has been discussed at great length with respect to on and off-policy RL training algorithms \citep{mnih2013playing}.
Other works cast optimizer meta-learning as RL \citep{li2016learning} for this very reason, at a large cost in terms of gradient variance.
In this work, 
we partially address this issue by training a large number of workers in parallel, to always maintain a diverse set of states when computing gradients.

For similar reasons, the number of steps per truncation, and the total number of unsupervised training steps, are both sampled 
in an attempt to limit the 
bias introduced by truncation. 

We found restricting the maximum inner loop step size to be crucial for stability.  \citet{pearlmutter1996investigation} studied the effect of learning rates with respect to the stability of optimization and showed that as the learning rate increases gradients become chaotic. This effect was later demonstrated with respect to neural network training in \citet{maclaurin2015gradient}. 
If learning rates are not constrained, we found that they rapidly grew and entered this chaotic regime.

Another technique we found useful in addressing these problems
is the use of batch norm in both the base model and in the $\text{UnsupervisedUpdate}$ rule. Multi-layer perceptron training traditionally requires very precise weight initialization for learning to occur. 
Poorly scaled initialization can make learning impossible \citep{schoenholz2016deep}. 
When applying a learned optimizer, especially early in meta-training of the learned optimizer, it is very easy for the learned optimizer to cause high variance weights in the base model, after which recovery is difficult. Batch norm helps solve this issues by making more of the weight space usable.

\section{Distributed Implementation}
\label{sec distributed implementation}

We implement the described models in distributed Tensorflow \citep{abadi2016tensorflow}. We construct a cluster of 512 workers, each of which computes gradients of the meta-objective asynchronously. Each worker trains on one task by first sampling a dataset, architecture, and a number of training steps. Next, each worker samples $k$ unrolling steps, does $k$ applications of the $\text{UnsupervisedUpdate}(\cdot; \theta)$, computes the $\text{MetaObjective}$ on each new state, computes $\pd{[\text{MetaObjective}]}{\theta}$ and sends this gradient to a parameter server. The final base-model state, $\phi$, is then used as the starting point for the next unroll until the specified number of steps is reached. These gradients from different workers are batched and $\theta$ is updated with asynchronous SGD. By batching gradients as workers complete unrolls, we eliminate most gradient staleness while retaining the compute efficiency of asynchronous workers, especially given heterogeneous workloads which arise from dataset and model size variation. An overview of our training can be seen in algorithm \ref{app:algorithm}. Due to the small base models and the sequential nature of our compute workloads, we use multi core CPUs as opposed to GPUs. Training occurs over the course of $\sim$8 days with $\sim$200 thousand updates to $\theta$ with minibatch size 256.

\newpage
\section{More Filters Over Meta-Training} \label{app:filters}
\begin{figure}[h!]
\centering
\includegraphics[width=0.99\textwidth]{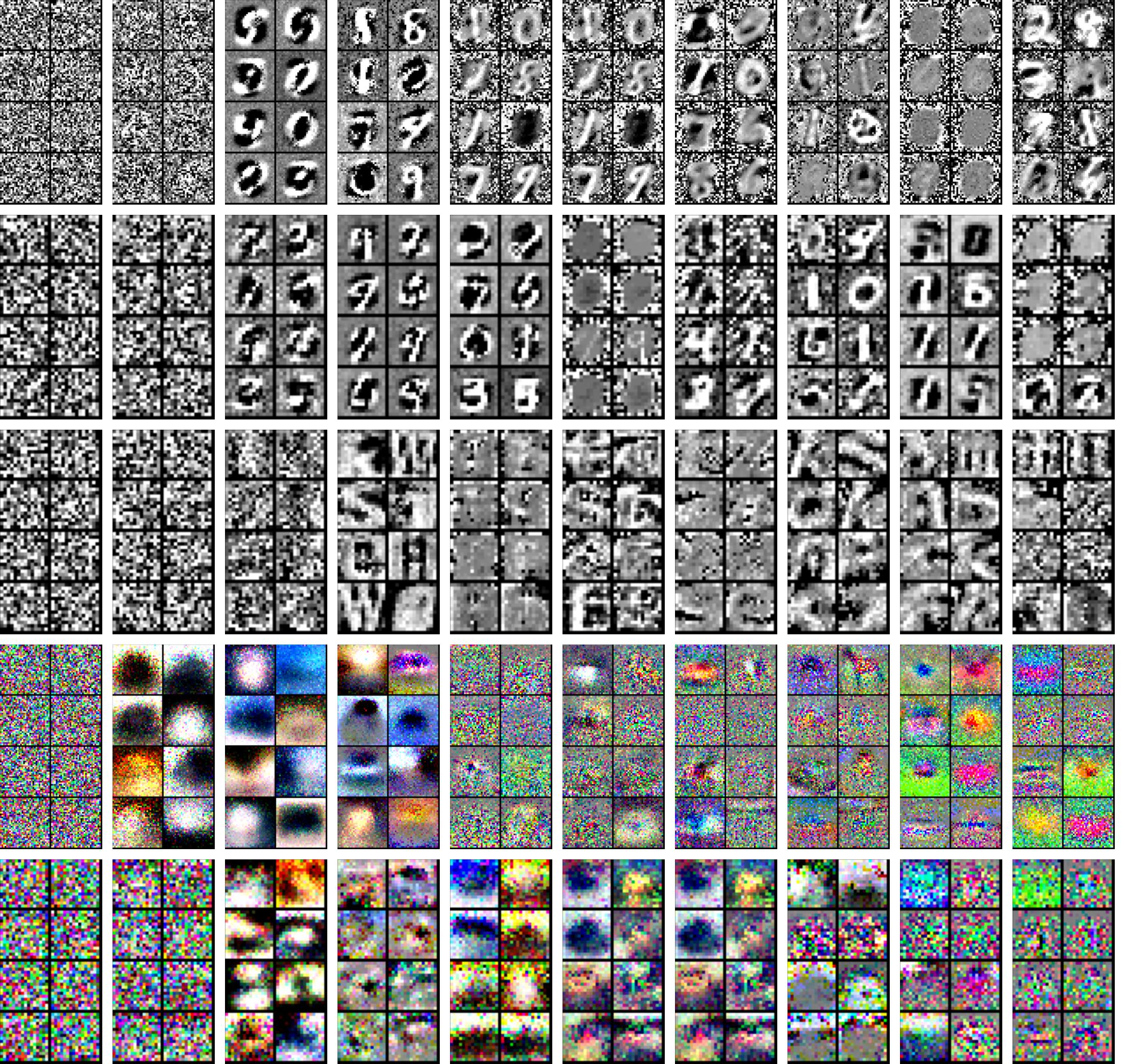}
\caption{ More filters extracted over the course of meta-training. Note, due to implementation reasons, the columns do not represent the same point / same iteration of $\theta$. Each filter is extracted after 10k inner loop optimization steps, at $\phi_{10k}$. From top to bottom we show: MNIST, Tiny MNIST, Alphabet, CIFAR10, and Mini CIFAR10. Filters shift from noise at initialization to more local features later in meta-training.
}
\end{figure}

\newpage
\section{Meta-Training curves} \label{app:meta_optimization}
\begin{figure}[h!]
\centering
\includegraphics[width=0.99\textwidth]{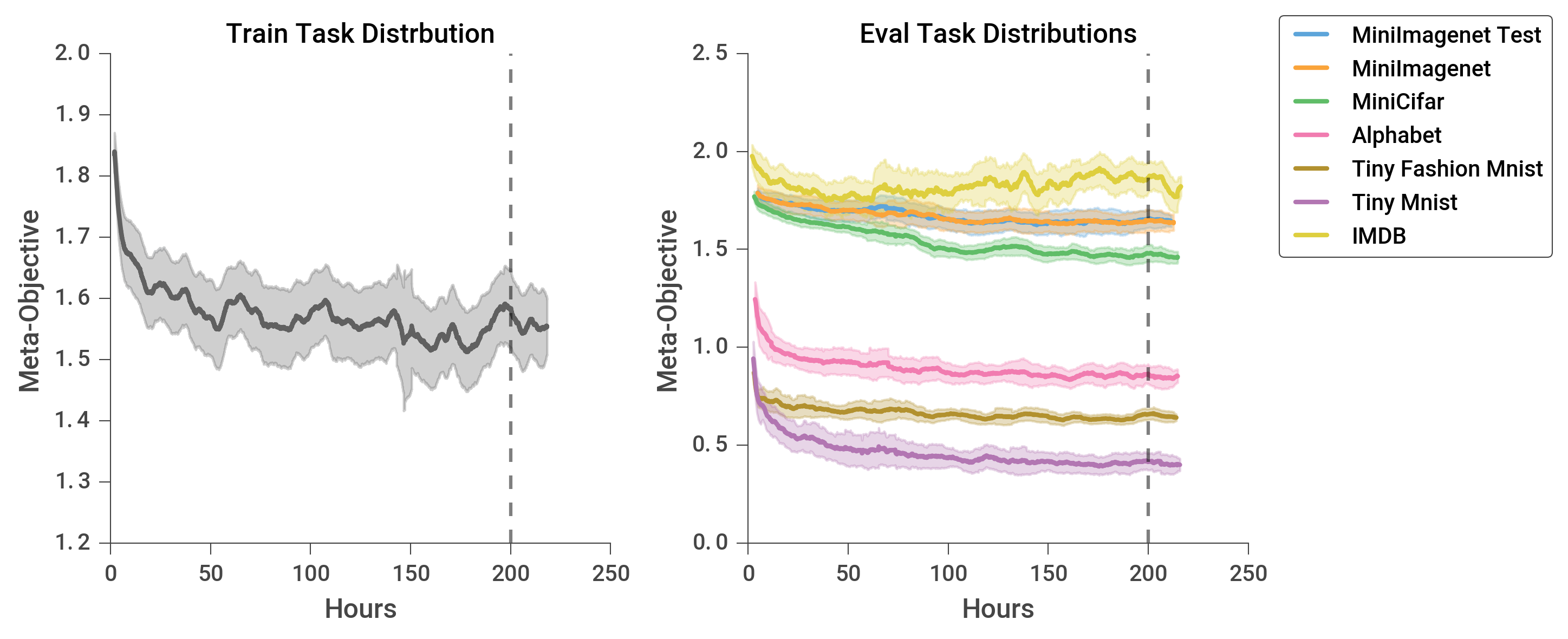}
\caption{Training curves for the meta-training and meta-evaluation task distributions. Our meta-train set consists of Mini Imagenet, Alphabet, and MiniCIFAR10. Our meta-test sets are Mini Imagenet Test, Tiny Fashion MNIST, Tiny MNIST and IMDB. Error bars denote standard deviation of evaluations with a fixed window of samples evaluated from a single model.
}
\end{figure}

\section{Learning algorithm} \label{app:algorithm}
\begin{algorithm}[H]
  \caption{Distributed Training Algorithm}
  \label{alg:example}
\begin{algorithmic}
  \STATE Initialize $\text{UnsupervisedUpdate}$s parameters, $\theta_0$.
  \STATE Initialize meta-training step count $v \leftarrow 0$.
  \STATE Initialize shared gradient state $\mathcal{S}$
  \STATE Initialize $R$ to be max meta-training steps.
  \WHILE{$r$ < $R$ on 512 workers in parallel}
      \STATE Sample supervised task, $\mathcal{D}$
      \STATE Sample base model $f(\cdot;\phi)$ and initialize $\phi_0$ randomly
      \STATE Sample $K$ truncated iterations
      \STATE Initialize learner iteration count: $t \leftarrow 0$
      \FOR{$k=1$ {\bfseries to} $K$}
            \STATE Sample $U$ unroll iterations
            \FOR{$u=0$ {\bfseries to} $U$}
              \STATE Sample data, $x$, from $\mathcal{D}$
              \STATE $\phi_{t+u+1} = \text{UnsupervisedUpdate}(x, \phi_{t+u}; \theta_r)$
            \ENDFOR
            \STATE Initialize MetaObjective accumulator $J \leftarrow 0$
            \FOR{$u=1$ {\bfseries to} $U$}
                \FOR{$m=1$ {\bfseries to} $M$ (number of $\text{MetaObjective}$ evaluations per iteration)}
                    \STATE Sample 2 batches of data, $x$,$x'$, and labels, $y$,$y'$ from $\mathcal{D}$ for both training and testing.
                    \STATE $J = J + \frac{1}{UM} \text{MetaObjective}(x, y, x', y', \phi_{t+u})$
                \ENDFOR
            \ENDFOR
            \STATE Compute $\pd{J}{\theta_r}$ for the last $U$ steps and store in $\mathcal{S}$
            \STATE Update current unroll iteration: $t \leftarrow t+U$
            \IF{size of $\mathcal{S} >$ meta-batch-size}
                \STATE Take averaged batch of gradients $G$ from $\mathcal{S}$
                \STATE $\theta_{r+1} = \theta_r - \operatorname{AdamUpdate}(G)$
                \STATE $r \leftarrow r+1$
            \ENDIF
      \ENDFOR
  \ENDWHILE
\end{algorithmic}
\end{algorithm}

\section{Model specification}
\label{app:modelspec}
In this section we describe the details of our base model and learned optimizer. First, we describe the inner loop, and then we describe the meta-training procedure in more depth.

In the following sections, $\lambda_{name}$ denotes a hyper-parameter for the learned update rule.

The design goals of this system are stated in the text body. 
At the time of writing, we were unaware of any similar systems, so as such the space of possible architectures was massive. Future work consists of simplifying and improving abstractions around algorithmic components.

An open source implementation of the $\text{UnsupervisedUpdate}$ can be found at \url{https://github.com/tensorflow/models/tree/master/research/learning_unsupervised_learning}.

\subsection{Inner Loop}

The inner loop computation consists of iterative application of the $\text{UnsupervisedUpdate}$ on a Base Model parameterized by $\phi$,
\begin{align}
    \phi_{t+1} = \text{UnsupervisedUpdate}(\cdot, \phi_t; \theta)
    ,
\end{align}
where $\phi$ consists of forward weights, $W^l$, biases, $b^l$, parameterizing a multi-layer perceptron as well as backward weights, $V^l$ used by $\text{UnsupervisedUpdate}$ when inner-loop training.

This computation can be broken down further as a forward pass on an unlabeled batch of data,
\begin{align}
    x^0 &\sim \mathcal{D} \\
    \{x^1..x^L, z^1..z^L\} &= f(x^0; \phi_t)
    ,
\end{align}
where $z^l$, and $x^l$ are the pre- and post-activations on layer $l$, and $x^0$ is input data.
We then compute weight updates:
\begin{align}
    \{(\Delta{W^{1\cdots L}})_t, (\Delta{b^{1\cdots L}})_t, (\Delta{V^{1\cdots L}})_t\} &= \operatorname{ComputeDeltaWeight}(x^0\cdots x^L, z^1\cdots z^L, \phi_t; \theta)
\end{align}

Finally, the next set of $\phi$ forward weights ($W^l$), biases ($b^l$), and backward weights ($V^l$), are computed. We use an SGD like update but with the addition of a decay term. Equivalently, this can be seen as setting the weights and biases to an exponential moving average of the $\Delta{W^l}$, $\Delta{V^l}$, and $\Delta{b^l}$ terms.
\begin{align}
    W^l_{t+1} &= W^l_t (1-\lambda_{\phi lr}) + \Delta{W^l} \lambda_{\phi lr}  \\
    V^l_{t+1} &= V^l_t (1-\lambda_{\phi lr}) + \Delta{V^l} \lambda_{\phi lr}  \\
    b^l_{t+1} &= b^l_t (1-\lambda_{\phi lr}) + \Delta{b^l} \lambda_{\phi lr} 
\end{align}

We use $\lambda_{\phi lr} = 3e-4$ in our work. $\Delta{W^l}$, $\Delta{V^l}$, $\Delta{b^l}$ are computed via meta-learned functions (parameterized by $\theta$).

In the following sections, we describe the functional form of the base model, $f$, as well as the functional form of $\operatorname{ComputeDeltaWeight}\left(\cdot;\theta\right)$.

\subsection{Base Model}
The base model, the model our learned update rule is training, is an $L$ layer multi layer perception with batch norm. We define $\phi$ as this model's parameters, consisting of weights ($W^l$) and biases ($b^l$) as well as the backward weights ($V^l$) used only during inner-loop training (applications of $\text{UnsupervisedUpdate}$). We define $N^1..N^L$ to be the sizes of the layers of the base model and $N^0$ to be the size of the input data.

\begin{align}
    \phi &= \{W^1..W^L, V^1..V^L, b^1..b^L\},
\end{align}
where
\begin{align}
    W^l &\in \R{N^{l-1}, N^{l}}\\
    V^l &\in \R{N^{l-1}, N^{l}}\\
    b^l  &\in \R{N^{l}}
\end{align}
where $N^{l}$ is the hidden size of the network, $N^0$ is the input size of data, and $N^{L}$ is the size of the output embedding. In this work, we fix the output layer: $N^{L} = 32$ and vary the remaining the intermediate $N^{1..(L-1)}$ hidden sizes.

The forward computation parameterized by $\phi$, consumes batches of unlabeled data from a dataset $\mathcal{D}$:
\begin{align}
    x^0 &\sim \mathcal{D}, x^0 \in \R{B, N^{0}} \\
    z^l &= \operatorname{BatchNorm}\left(x^{l-1}W^l\right) + b^l \\
    x^l &= \operatorname{ReLU(z^l)}
\end{align}
for $l=1..L$.

We define $f(x; \phi)$ as a function that returns the set of internal pre- and post-activation function hidden states as well as $\hat{f}(x; \phi)$ as the function returning the final hidden state:
\begin{align}
    f(x, \phi) &= \{z^1...z^L, x^0..x^L, \phi\} \\
    \hat{f}(x, \phi) &= x^L
\end{align}

\subsection{$\text{MetaObjective}$}\label{app:detail:MetaObjective}

We define the $\text{MetaObjective}$ to be a few shot linear regression. To increase stability and avoid undesirable loss landscapes, we additionally center, as well as normalize the predicted target before doing the loss computation. The full computation is as follows:
\begin{align}
    \text{MetaObjective}(x, y, x', y', \phi) :: (\R{B, N^{0}}, \R{B, N^{classes}}, \R{B, N^{0}}, \R{B,N^{classes}}, \Phi) \rightarrow \R{1},
\end{align}
where $N^{classes}$ is the number of classes, and $y$, $y'$ are one hot encoded labels.

First, the inputs are converted to embeddings with the base model,
\begin{align}
    x^L &= \hat{f}(x; \phi) \\   
    {x'^L} &= \hat{f}({x'}; \phi) 
\end{align}
Next, we center and normalize the prediction targets. We show this for $y$, but $y'$ is processed identically.
\begin{align}
    \bar{y} &= \frac{1}{BN^{classes}}\sum_{i}^{B}\sum_{j}^{N^{classes}}{y_{ij}} \\
    \hat{y}_{ij}  &= \dfrac{y_{ij} - \bar{y}}{
        \sqrt{
            \frac{1}{N^{classes}}\sum_{a}^{N^{classes}}{\left\| \bar{y}_{ia} \right\|_2^2}
            }}
\end{align}

We then solve for the linear regression weights in closed form with features: $x^L$ and targets: $\hat{y}$. We account for a bias by concatenating a $1$'s vector to the features.
\begin{align}
    A &= [x^L; \mathbbm{1}] \\
    C &= \left((A^tA) + I \lambda_\text{ridge}\right)^{-1}A^T\hat{y}
\end{align}

We then use these inferred regression weights $C$ to make a prediction on the second batch of data, normalize the resulting prediction, and compute a final loss,
\begin{align}
    p &= C[x'^L; \mathbbm{1}] \\
    \hat{p}_{bi} &= \dfrac{p_{bi}}{\left\| p_{b} \right\|_2} \\
    \text{MetaObjective}(\cdot) &= \frac{1}{B}\sum_{b}^B{\left\| \hat{p}_{b} - \hat{y}_{b} \right\|_2^2}
    .
\end{align}
Note that due to the normalization of predictions and targets, this corresponds to a cosine distance (up to an offset and multiplicative factor).

\subsection{$\text{UnsupervisedUpdate}$}
\label{app:unsupervisedupdate_sub}
The learned update rule is parameterized by $\theta$. In the following section, we denote all instances of $\theta$ with a subscript to be separate named learnable parameters of the $\text{UnsupervisedUpdate}$. $\theta$ is shared across all instantiations of the unsupervised learning rule (shared across layers). It is this weight sharing that lets us generalize across different network architectures.

The computation is split into a few main components: first, there is the forward pass, defined in $f(x; \phi)$. Next, there is a ``backward'' pass, which operates on the hidden states in a reverse order to propagate an error signal $\delta^l$ back down the network. In this process, a hidden state $h^l$ is created for each layer. These h are tensors with a batch, neuron and feature index: $h^l \in \R{B,N^l, \lambda_{hdims}}$ where $\lambda_{hdims}=64$. Weight updates to $W^l$ and $b^l$ are then readout from these $h^l$ and other signals found both locally, and in aggregate along both batch and neuron.

\subsubsection{Backward error propagation}
In backprop, there exists a single scalar error signal that gets backpropagated back down the network. In our work, this error signal does not exist as there is no loss being optimized. Instead we have a learned top-down signal, $d^L$, at the top of the network. Because we are no longer restricted to the form of backprop, we make this quantity a vector for each unit in the network, rather than a scalar,
\begin{align}
    x^0 &\sim \mathcal{D}, x^0 \in \R{B, N_{x}} \\
    \{z^1...z^L, x^0..x^L, \phi\} &= f(x; \phi) \\
    d^L &= \operatorname{TopD}(x^L; \theta_{topD}) \\
\end{align}
where $d^l \in \R{B, N^l, \lambda_{deltadims}}$. In this work we set $\lambda_{deltadims}=32$. The architecture of $\operatorname{TopD}$ is a neural network that operates along every dimension of $x^L$. The specification can be found in \ref{sec topD}.

We structure our error propagation 
similarly to 
the structure of the backpropagated error signal in a standard MLP with contributions to the error at every layer in the network,
\begin{align}
  \delta^{l}_{ijd} &= d^{l}_{ijd} \odot \sigma(z^{l}_{ijd}) + \sum_k^{\lambda_{hdims}} \left(\theta_{errorPropW}\right)_{kd}h_{ijk}^{l} + \left(\theta_{errorPropB}\right)_d
\end{align}
where $\delta^l$ has the same shape as $d^l$, and $\theta_{errorPropW} \in \R{{\lambda_{hdims}},{\lambda_{deltadims}}}$ and $\theta_{errorPropB} \in \R{\lambda_{deltadims}}$.
Note both $\theta_{errorPropW}$ and $\theta_{errorPropB}$ are shared for all layers $l$. 

In a similar way to backprop, we move this signal down the network via multiplying by a backward weight matrix ($V^l$). We do not use the previous weight matrix transpose as done in backprop, instead we learn a separate set of weights that are not tied to the forward weights and updated along with the forward weights as described in \ref{sec app weight updates}. Additionally, we normalize the signal to have fixed second moment,
\begin{align}
    \tilde{d}^{l}_{imd} &= \sum_j^{N^{l+1}} \delta_{ijd}^{l+1} (V^{l+1})_{mj} \\
    d^{l}_{imd} &= \hat{d}^{l}_{imd} \left(
    \frac{1}{\lambda_{deltadims}}
    \sum_a^{\lambda_{deltadims}} \tilde{d}^l_{ima}
    \right)^{-\frac{1}{2}}
\end{align}

The internal $h^l \in \R{B, N^l, \lambda_{hdims}}$ vectors are computed via:
\begin{align}
    h^l = \operatorname{ComputeH}\left(d^l, x^l, z^l; \theta_{computeH} \right)
\end{align}

The architecture of $\operatorname{ComputeH}$ is a neural network that operates on every dimension of all the inputs. It can be found in \ref{sec computeH}. These definitions are recursive, and are computed in order: $h^L,h^{L-1}\cdots h^1,h^0$. With these computed, weight updates can be read out (Section \ref{sec app weight updates}). When the corresponding symbols are not defined (e.g. $z^0$) a zeros tensor with the correct shape is used instead.

\subsection{Weight Updates}
\label{sec app weight updates}
The following is the implementation of $\operatorname{ComputeDeltaWeight}(x^0\cdots x^L, z^1\cdots z^L, \phi; \theta)$

For a given layer, $l$, our weight updates are a mixture of multiple low rank readouts from $h^l$ and $h^{l-1}$. These terms are then added together with a learnable weight, in $\theta$ to form a final update. The final update is then normalized mixed with the previous weights. We update both the forward, $W^l$, and  the backward, $V^l$, using the same update rule parameters $\theta$. We show the forward weight update rule here, and drop the backward for brevity.

For convenience, we define a low rank readout function $\lowRR$ that takes $h^l$ like tensors, and outputs a single lower rank tensor.
\begin{align}
    \lowRR&\left(h^a, h^b ; \Theta \right) :: \\
    &\left( \R{B, N^a, \lambda_{hdims}}, \R{B, N^b, \lambda_{hdims}} \right) \rightarrow \R{N"^a, N^b}
\end{align}
Here, $\Theta$, is a placehoder for the parameters of the given readout. 
$\lowRR$ is defined as:
\begin{align}
    \Theta &= \{P^a, P^b\} \quad \text{where} \enskip P^a \in \R{\lambda_{hdims}, \lambda_{gradc}} \enskip \text{and} \enskip P^b \in \R{\lambda_{hdims}, \lambda_{gradc}}\\
    r^a_{ijp} &= \sum_k h^a_{ijk} P^a_{kp} \\
    r^b_{ijp} &= \sum_k h^b_{ijk} P^b_{kp} \\
    \lowRR\left( \cdot \right)_{jp} &= \sum^B_i \sum^{\lambda_{gradc}}_{k} r^b_{ijk} r^a_{ipk} \frac{1}{B \lambda_{hdims}}    
\end{align}
where $\lambda_{gradc}=4$ and is the rank of the readout matrix (per batch element).

\subsubsection{Local terms}
This sequence of terms allow the weight to be adjusted as a function of state in the pre- and post-synaptic neurons. They should be viewed as a basis function representation of the way in which the weight changes as a function of pre- and post-synaptic neurons, and the current weight value. We express each of these contributions to the weight update as a sequence of weight update planes, with the $i$th plane written $\Delta W^l_i \in \R{N^{l-1}\times N^l}$.
Each of these planes will be linearly summed, with coefficients generated as described in Equation \ref{eq:planeavg}, in order to generate the eventual weight update.

\begin{align}
    \hat{W}^l &= \frac{W^l}{\sqrt{\frac{1}{N^{l-1}}\sum_{i}^{N^{l-1}}{(W^l)_i^2}}} \\ 
    \Delta W_1^l &= \hat{W}^l \\
    \Delta W_2^l &= (\hat{W}^l)^2 sign(\hat{W}^l)  \\
    \Delta W_3^l &= \lowRR(h^{l-1}, h^{l}; \theta_{zero}) \\
    \Delta W_4^l &= \operatorname{exp}(-(\hat{W}^l_b)^2) \odot \lowRR(h^{l-1}, h^{l}; \theta_{rbf}) \\
    \Delta W_5^l &= W^l_b \odot \lowRR(h^{l-1}, h^{l}; \theta_{first}) \\
    \Delta W_6^l &= \frac{1}{B}\sum_{b}^{B}{
        \left(x^{l-1}_b - \frac{1}{B}\sum_{b'}^{B}{x^{l-1}_{b'}}\right)^T
        \left(x^{l}_b - \frac{1}{B}\sum_{b'}^{B}{x^{l}_{b'}}\right)^T
        }
\end{align}

\subsubsection{Decorrelation terms}
Additional weight update planes are designed to aid units in remaining decorrelated from each other's activity, and in decorrelating their receptive fields. Without terms like this, a common failure mode is for many units in a layer to develop near-identical representations. Here, $S^l_i$ indicates a scratch matrix associated with weight update plane $i$ and layer $l$.
\begin{align}
    S^l_7 &= \frac{1}{\sqrt{N^{l-1}}} \lowRR(h^{l-1}, h^{l-1}, \theta_{linLowerSymm}) \\
    \Delta W_7^l &=  \frac{1}{\sqrt{2}}\left[\left(S^l_7 + \pp{S^l_7}^T\right) \odot (1-I) \right] W^l \\
    S^l_8 &= \frac{1}{\sqrt{N^{l-1}}} \lowRR(h^{l-1}, h^{l-1}, \theta_{sqrLowerSymm}) \\
    \Delta W_8^l &=  \frac{1}{\sqrt{2}}\left[\left(S^l_8 + \pp{S^l_8}^T\right) \odot (1-I) \right]  \left(\sqrt{1+(W^l)^2}-1\right) \\
    S^l_9 &= \frac{1}{\sqrt{N^{l}}} \lowRR(h^{l}, h^{l}, \theta_{linUpperSymm}) \\
    \Delta W_9^l &=  \frac{1}{\sqrt{2}} W^l \left[\left(S^l_9 + \pp{S^l_9}^T\right) \odot (1-I) \right] \\
    S^l_{10} &= \frac{1}{\sqrt{N^{l}}} \lowRR(h^{l}, h^{l}, \theta_{sqrUpperSymm}) \\
    \Delta W_{10}^l &=  \frac{1}{\sqrt{2}}\left(\sqrt{1+(W^l)^2}-1\right)\left[\left(S^l_{10} + \pp{S^l_{10}}^T\right) \odot (1-I) \right] \\
\end{align}

\subsection{Application of the weight terms in the optimizer}
We then normalize, re-weight, and merge each of these weight update planes into a single term, which will be used in the weight update step,
\begin{align}
    \label{eq:planeavg}
    \tilde{\Delta W_i^l} &= \frac{\Delta W_i^l}{
        \sqrt{1 + \frac{1}{N^{l-1}N^{l}}\sum_{m}^{N^{l-1}}\sum_{n}^{N^{l}}{\left(\Delta W_{imn}^l\right)^2}}
    } \\
    (\Delta {W^l_{merge}})_{jk} &= \frac{1}{B}\sum_{i}^{B}{(\theta_{mergeW})_{i} \tilde{\Delta W_{ijk}^l}},
\end{align}

where $\theta_{mergeW} \in \R{10}$ (as we have 10 input planes).

To prevent pathologies during training, we perform two post processing steps to prevent the learned optimizer from cheating, and increasing its effective learning rate, leading to instability. We only allow updates which do not decrease the weight matrix magnitude, 
\begin{align}
    \Delta {W^l_{orth}} &= \Delta {W^l_{merge}} - \hat{W}^l \ReLU(\Delta {W_{merge}} \cdot \hat{W}^l),
\end{align}

where $\hat{W}^l$ is ${W}^l$ scaled to have unit norm, 
and we normalize the length of the update,
\begin{align}
    \Delta {W^l_{final}} &= 
    \frac{\Delta {W^l_{orth}}}
    {
        \sqrt{1 + \frac{1}{N^{l-1}N^{l}}\sum_{m}^{N^{l-1}}\sum_{n}^{N^l}{\left(\Delta {W^l_{orth}}\right)_{mn}^2}}
    }
\end{align}

To compute changes in the biases, we do a readout from $h^l$. We put some constraints on this update to prevent the biases from pushing all units into the linear regime, and minimizing learning. We found this to be a possible pathology.
\begin{align}
    \Delta b^l_{base} &= \frac{1}{B} \sum_i^B \sum_k^{\lambda_{hdims}} (\theta_{Breadout})_k h^{l}_{ijk} \\
    \Delta b^l_{constrained} &= \Delta b^l_{base} - \ReLU\left( - \frac{1}{N^l}\sum_{i}^{N^l}{\left(\Delta b^l_{base}\right)_i} \right),
\end{align}
where $\theta_{Breadout} \in \R{\lambda_{hdims}}$.

We then normalize the update via the second moment:
\begin{align}
    \Delta b^l_{final} &= \frac{\Delta b^l_{constrained}}{\frac{1}{N^l}\sum_i^{N^l}(\Delta b^l_{constrained})^2_i}
\end{align}

Finally, we define $\operatorname{ComputeDeltaWeight}$ as all layer's forward weight updates, and backward weight updates, and bias updates.
\begin{align}
    \operatorname{ComputeDeltaWeight}(x^0\cdots x^L, z^1\cdots z^L, \phi_t; \theta) = (\Delta W^{1..L}_{final}, \Delta b^{1..L}_{final}, \Delta V^{1..L}_{final})
\end{align}

\subsection{$\operatorname{TopD}$}
\label{sec topD}
This function performs various convolutions over the batch dimension and data dimension. For ease of notation, we use $m$ as an intermediate variable. Additionally, we drop all convolution and batch norm parameterizations. They are all separate elements of $\theta_{topD}$. We define two 1D convolution operators that act on rank 3 tensors: $\ConvBatch$ which performs convolutions over the zeroth index, and $\ConvUnit$ which performs convolutions along the firs index. We define the $S$ argument to be size of hidden units, and the $K$ argument to be the kernel size of the 1D convolutions. Additionally, we set the second argument of $\BatchNorm$ to be the axis normalized over.

\begin{equation}
    \operatorname{TopD}\left(x^L; \theta_{topD} \right) :: \R{B, N^L} \rightarrow \R{B, N^L, \lambda_{deltadims}}
\end{equation}

First, we reshape to add another dimension to the end of $x^L$, in pseudocode:
\begin{align}
    m_0 &= [x^L]
\end{align}
Next, a convolution is performed on the batch dimension with a batch norm and a ReLU non linearity. This starts to pull information from around the batch into these channels.
\begin{align}
    m_1 &= \ConvBatch\left(m_0, S=\lambda_{topdeltasize}, K=5\right) \\
    m_2 &= \ReLU\left(\BatchNorm\left(m_1, [0,1]\right)\right) \\
\end{align}

We set $\lambda_{topdeltasize} = 64$.
Next, a series of unit convolutions (convolutions over the last dimension) are performed. These act as compute over information composed from the nearby elements of the batch. These unit dimensions effectively rewrite the batch dimension. This restricts the model to operate on a fixed size batch.
\begin{align}
    m_3 &= \ConvUnit\left(m_2, S=B, K=3\right) \\
    m_4 &= \ReLU\left(\BatchNorm\left(m_3, [0,1]\right)\right)\\
    m_5 &= \ConvUnit\left(m_4, S=B, K=3\right) \\
    m_6 &= \ReLU\left(\BatchNorm\left(m_5, [0,1]\right)\right)
\end{align}
Next a series of 1D convolutions are performed over the batch dimension for more compute capacity.
\begin{align}
    m_7 &= \ConvBatch\left(m_6, S=\lambda_{topdeltasize}, K=3\right) \\
    m_8 &= \ReLU\left(\BatchNorm\left(m_7, [0,1]\right)\right) \\
    m_9 &= \ConvBatch\left(m_8, S=\lambda_{topdeltasize}, K=3\right) \\
    m_{10} &= \ReLU\left(\BatchNorm\left(m_9, [0,1]\right)\right)
\end{align}

Finally, we convert the representations to the desired dimensions and output.
\begin{align}
    m_{11} &= \ConvBatch\left(m_{10}, S=\lambda_{deltadims}, K=3\right) \\
    \operatorname{TopD}\left(x^L; \theta_{topD}\right) &= m_{11}
\end{align}

\subsection{$\operatorname{ComputeH}$}
\label{sec computeH}
This is the main computation performed while transfering signals down the network and the output is directly used for weight updates. It is defined as:
\begin{multline}
\operatorname{ComputeH}\left(d^l, x^l, z^l, W^{l}, W^{l+1}, b^l; \theta_{computeH} \right) ::\\
\left(\R{B, N^l, \lambda_{deltadims}}, \R{B, N^l}, \R{B, N^l}, \R{N^{l-1}, N^{l}}, \R{N^{1}, N^{l+1}}, \R{N^l} \right) \rightarrow \left(\R{B, N^l, \lambda_{hdims}}\right)
\end{multline}

The outputs of the base model, ($x^l$, $z^l$), plus an additional positional embeddings are stacked then concatenated with $d$, to form a tensor in $\R{B, N^l, (4+\lambda_{deltadims})}$:
\begin{align}
    (p^0)_{ij} &= \operatorname{sin}(\dfrac{2j\pi}{N^l}) \\
    (p^1)_{ij} &= \operatorname{cos}(\dfrac{2j\pi}{N^l}) \\
    m_0 &= [x^l, z^l, p^0, p^1] \enskip \text{where} \enskip m_0 \in \R{B, N^l, 4}\\
    m_1 &= [m_0; d^l]
\end{align}

Statistics across the batch and unit dimensions, 0 and 1, are computed.
We define a $\operatorname{Stats_i}$ function bellow. We have 2 instances for both the zeroth and the first index, shown bellow is the zeroth index and the first is omitted.
\begin{align}
    \operatorname{Stats_0}\left(w\right) :: \R{K^0, K^1} \rightarrow \R{K^1, 4} \\
    (s_{l1})_j &= \frac{1}{K^0} \sum_i^{K^0} \operatorname{abs}(w_{ij}) \\
    (s_{l2})_j &= \sqrt{\frac{1}{K^0} \sum_i^{K^0} (w_{ij})^2} \\
    (s_{\mu})_j &= \frac{1}{K^0} \sum_i^{K^0} w_{ij} \\
    (s_{\sigma})_j &= \sqrt{\frac{1}{K^0} \sum_i^{K^0} ((s_{\mu})_i - w_{ij})^2} \\
    Stats_0\left(w\right) = [s_{l1}, s_{l2}, s_{\mu}, s_{\sigma}]
\end{align}

We the compute statistics of the weight matrix below, and above. We tile the statistics to the appropriate dimensions and concatenate with normalized inputs as well as with the bias (also tiled appropriately).
\begin{align}
    (s_0)_{ijk} &= (\operatorname{Stats_0}\left(W^{l}, 0 \right) )_j \\
    (s_1)_{ijk} &= (\operatorname{Stats_1}\left(W^{l+1}, 1 \right) )_i \\
    m_2 &= \BatchNorm(m_1, [0,1]) \\
    \hat{b}_{ijk} &= b^l_{j} \enskip \text{where} \enskip \hat{b} \in \R{B, N^l, 1}\\
    m_3 &= [s_0; s_1; m_2; \hat{b}] \enskip \text{where} \enskip m_3 \in \R{B, N^l, 4+4+(4+\lambda_{deltadims})+1}
\end{align}
With the inputs prepared, we next perform a series of convolutions on batch and unit dimensions, (0, 1).
\begin{align}
    m_4 &= \ConvBatch\left(m_3, S=\lambda_{computehsize}, K=3\right) \\
    m_5 &= \ReLU\left(\BatchNorm\left(m_4, [0,1]\right)\right) \\
    m_6 &= \ConvUnit\left(m_5, S=\lambda_{computehsize}, K=3\right) \\
    m_7 &= \ReLU\left(\BatchNorm\left(m_6, [0,1]\right)\right) \\
    m_8 &=\ConvBatch\left(m_7, S=\lambda_{computehsize}, K=3\right) \\
    m_9 &= \ReLU\left(\BatchNorm\left(m_8, [0,1]\right)\right) \\
    m_{10} &= \ConvUnit\left(m_9, S=\lambda_{computehsize}, K=3\right) \\
    m_{11} &= \ReLU\left(\BatchNorm\left(m_{10}, [0,1]\right)\right)
\end{align}
The result is then output.
\begin{align}
    \operatorname{ComputeH}\left(\cdot; \theta_{computeH} \right) = m_{11}
\end{align}
We set $\lambda_{computehsize} = 64$ which is the inner computation size. 

\section{Experimental details} \label{app:experimental}
\subsection{Meta Training}
\subsubsection{Training Data Distribution} \label{app:experimental:data}
We trained on a data distribution consisting of tasks sampled uniformly over the following datasets. Half of our training tasks where constructed off of a dataset consisting of 1000 font rendered characters. We resized these to 14x14 black and white images. We call this the glyph dataset. "Alphabet" is an example of such a dataset consisting of alphabet characters. We used a mixture 10, 13, 14, 17, 20, and 30 way classification problems randomly sampled, as well as sampling from three 10-way classification problems sampled from specific types of images: letters of the alphabet, math symbols, and currency symbols. For half of the random sampling and all of the specialized selection we apply additional augmentation. This augmentation consists of random rotations (up to 360 degrees) and shifts up to +-5 pixels in the x and y directions. The parameters of the augmentation were inserted into the regression target of the $\text{MetaObjective}$ as a curriculum of sorts and to provide diverse training signal.

In addition to the the glyph set, we additionally used Cifar10, resized to 16x16, as well as 10, 15, 20, and 25 way classification problems from imagenet. Once again we resized to 16x16 for compute reasons.

With a dataset selected, we apply additional augmentation with some probability consisting of the following augmentations. A per task dropout mask (fixed mask across all images in that task). A per example dropout mask (a random mask per image). A permutation sampled from a fixed number of pre created permutations per class. A per image random shift in the x direction each of image. All of these additional augmentations help with larger domain transfer.

\subsection{Meta-Optimization}
We employ Adam \citep{kingma2014adam} as our meta-optimizer. We use a learning rate schedule of 3e-4 for the first 100k steps, then 1e-4 for next 50k steps, then 2e-5 for remainder of meta-training. We use gradient clipping of norm 5 on minibatchs of size 256.

We compute our meta-objective by averaging 5 evaluation of the linear regression. We use a ridge penalty of 0.1 for all this work.

When computing truncated gradients, we initially sample the number of unrolled applications of the $\text{UnsupervisedUpdate}$ in a uniform distribution of [2,4]. This is the number of steps gradients are backpropogated through. Over the course of 50k meta-training steps we uniformly increase this to [8,15]. This increases meta-training speed and stability as large unrolls early in training can be unstable and don't seem to provide any value.

For sampling the number of truncated steps (number of times the above unrolls are performed), we use a shifted normal distribution -- a normal distribution with the same mean and standard deviation. We chose this based on the expected distribution of the training step, $\phi$ iteration number, across the cluster of workers. We initially set the standard deviation low, 20, but slowly increased it over the course of 5000 steps to 20k steps. This slow increase also improved stability and training speed.

\subsection{Experimental setup}
For each experimental figure, we document the details.

\subsubsection{Objective Function Mismatch} \label{app:vae_exp}
The VAE we used consists of 3 layers, size 128, with ReLU activations and batch norm between each layer. We then learn a projection to mean and log std of size 32. We sample, and use the inverse architecture to decode back to images. We use a quantized normal distribution (once again parameterized as mean and log std) as a posterior. We train with Adam with a learning rate of 1e-4. To isolate the effects of objective function mismatch and overfitting, we both train on the unlabeled training set and evaluate on the labeled training set instead of a validation set.

\subsubsection{Generalization: Dataset and Domain}
We use a 4 layer, size 128 unit architecture with a 32 layer embedding for all models. We select performance at 100k training steps for the VAE, and 3k for our learned optimizer.

Our supervised learning baseline consists of the same architecture for the base model but with an additional layer that outputs log probabilities. We train with cross entropy loss on a dataset consisting of only 10 examples per class (to match the other numbers). Surprisingly, the noise from batch norm acts as a regularizer, and allowing us to avoid needing a complex early stopping scheme as test set performance simply plateaus over the course of 10k steps. We train with Adam with a learning rate of 3e-3, selected via a grid over learning rate on test set performance. In this setting, having a true validation set would dramatically lower the amount of labeled data available (only 100 labeled examples) and using the test set only aids in the this baseline's performance.

For the IMDB experiments, we tokenized and selected the top 1K words with an additional set of tokens for unknown, sentence start, and sentence end. Our encoding consisted of a 1 if the word is present, otherwise 0. We used the same 4 layer, 128 hidden unit MLP with an addition layer outputting a 32 dimensional embedding.

\subsubsection{Generalization: Network Architecture}
We used ReLU activations and 4 layers of size 128 with an additional layer to 32 units unless otherwise specified by the specific experiment.

\subsubsection{Existing meta learning models}
We trained prototypical networks \citep{snell2017prototypical} on either intact or shuffled mini-Imagenet images. For shuffled images, we generated a fixed random permutation for the inputs independently for every instantiation of the base network (or 'episode' in the meta-learning literature \citep{Vinyals2016,ravi2016optimization}). The purpose of shuffling was to demonstrate the inductive bias of this type of meta-learning, namely that they do not generalize across data domain. Note that the base network trained was the same fully connected architecture like that used in this paper (3 layers, size 128, with ReLU activations and batch normalization between layers). Though the original paper used a convolutional architecture, here we swapped it with the fully connected architecture because the tied weights in a convolutional model do not make sense with shuffled pixels.

\end{document}